\documentclass[review,3p]{elsarticle}

\usepackage{amssymb}
\usepackage{mathtools,amsmath,amssymb} 
\usepackage{mathrsfs,amsfonts,dsfont}
\usepackage{tabularx,booktabs,multirow}
\usepackage{graphicx}
\usepackage{rotating}
\usepackage{multirow}
\usepackage{lscape}
\usepackage{bbm}
\usepackage{url}
\usepackage{lscape}
\usepackage{todonotes}

\usepackage{ulem}

\newcommand{\citeA}[1]{\citet{#1}}
\newcommand{\new}[1]{\textcolor{black}{#1}}

\journal{Information Sciences}

\begin{document}

\begin{frontmatter}



\title{Characterizing classification datasets: a study of meta-features for meta-learning}


\author[utfpr,icmc]{Adriano Rivolli}
\ead{rivolli@utfpr.edu.br}

\author[unb,icmc]{Lu\'is P. F. Garcia}
\ead{luis.garcia@unb.br}

\author[feup]{Carlos Soares}
\ead{csoares@fe.up.pt}

\author[tue]{Joaquin Vanschoren}
\ead{j.vanschoren@tue.nl}

\author[icmc]{Andr\'e C. P. L. F. de Carvalho}
\ead{andre@icmc.usp.br}

\address[utfpr]{Computing Department, Technological University of Paran\'a, \\Av. Alberto Carazzai, 1640, Corn\'elio Proc\'opio, Paran\'a 86300-000, Brazil}

\address[unb]{Department of Computer Science, University of Brasilia, \\Asa Norte, Bras\'ilia, Distrito Federal 70910-900, Brazil}

\address[icmc]{Institute of Mathematical and Computer Sciences, University of S\~ao Paulo, \\Av. Trabalhador S\~ao-carlense, 400, S\~ao Carlos, S\~ao Paulo 13560-970, Brazil}

\address[feup]{Faculty of Engineering, University of Porto\\ Rua Dr. Roberto Frias, s/n, 4200-465 Porto, Portugal}

\address[tue]{ Department of Mathematics and Computer Science, Eindhoven University of Technology\\ P.O.Box 513, 5600MB Eindhoven, Netherlands}
            
\begin{abstract}
Meta-learning is increasingly used to support the recommendation of machine learning algorithms and their configurations. 
Such recommendations are made based on meta-data, consisting of performance evaluations of algorithms on prior datasets, as well as
characterizations of these datasets. 
These characterizations, also called meta-features, describe properties of the data which are predictive for the performance of machine learning algorithms trained on them. 
Unfortunately, despite being used in a large number of studies, meta-features are not uniformly described, organized and computed, making many empirical studies irreproducible and hard to compare.
This paper aims to deal with this by systematizing and standardizing data characterization measures for classification datasets used in meta-learning.
Moreover, it presents MFE, a new tool for extracting meta-features from datasets and identifying more subtle reproducibility issues in the literature, proposing guidelines for data characterization that strengthen reproducible empirical research in meta-learning.
\end{abstract}




\end{frontmatter}


\section{Introduction}
\label{sec:introduction}

Machine learning algorithms have an inductive bias: they each make assumptions about the data distribution and choose specific generalization hypotheses over several other possible generalizations, thus restricting the search space \citep{Mitchell1997,Wolpert1992}. 
Since the true data distribution is unknown, several techniques are typically tried to achieve a satisfactory solution for a particular task. 
This trial-and-error approach is laborious and subjective, given the many choices that need to be made. 
Alternatively, meta-learning (MtL) presents a data-driven, automatic selection of techniques, by using knowledge extracted from previous tasks \citep{Brazdil2009}. 
For instance, a meta-model can be trained on prior tasks to recommend suitable techniques for a new task \citep{Vanschoren2012}.

Such a recommender system requires a systematic collection of dataset characteristics, along with the corresponding performance of different algorithms. 
These characteristics extracted from the datasets, named meta-features, play a crucial role in the successful use of MtL \citep{Bensusan2001,Bilalli2017}. 
Many empirical studies have investigated the effectiveness of meta-features in different domains \citep{Bensusan2000a,Bensusan2001,Filchenkov2015,Furnkranz2001,Peng2002,Pfahringer2000,Reif2011,Reif2014}, and 
proposed different sets of meta-features to characterize a given MtL task.

Unfortunately, several aspects that affect the reproducibility and generalizability of these experiments have been neglected or ignored in the literature. 
These include details concerning the dataset characterization process, hyperparameter settings used to evaluate algorithms, as well as procedures that deal with data encoding and missing values. 
These aspects require additional and careful investigation, especially given the current reproducibility crisis in machine learning research \citep{Hutson2018}. 

The lack of a systematic approach to compute meta-features has obfuscated the analyses in empirical MtL studies. 
To overcome this limitation, \citeA{Pinto2016} proposed a framework to systematize the extraction of meta-features, defining a meta-feature in terms of three components: \emph{meta-function}, \emph{object} and \emph{post-processing}.
In short, a \emph{meta-function} (e.g. entropy) extracts conceptual information from the \emph{object} (e.g. predictive attributes) and a \emph{post-processing} function (e.g. mean) summarizes the result.
Different variations of these three components result in different meta-features. 
The authors claim that all current meta-features can be decomposed using these three components. 
However, this framework does not directly mitigate the reproducibility problem, since the formalization, categorization and development of the meta-features are not addressed in the framework.

A good initiative to overcome this problem is OpenML \citep{OpenML2013}, an on-line research platform that supports a standard characterization of datasets. 
As such, OpenML allows the comparison of MtL studies, insofar as they use the meta-features computed by OpenML. 
This set of meta-features is itself not defined systematically, however, which may hamper their suitability for subsequent meta-learning studies.

This paper surveys a comprehensive list of meta-features and their usage in the data classification MtL literature, and systematically organizes and categorizes these meta-features in a taxonomy.
Furthermore, it highlights the main strengths and weaknesses of each meta-feature, identifying important reproducibility issues related to them.
Finally, the paper presents the Meta-Feature Extractor (MFE) tool to compute many of these meta-features. Publicly available as a package in Python\footnote{\url{https://pypi.org/project/pymfe/}} and in R\footnote{\url{https://CRAN.R-project.org/package=mfe}}, MFE offers a flexible and standalone implementation of meta-features for MtL experiments.


\section{Taxonomy} 
\label{sec:taxonomy}

Let $\mathcal{D}$ be a dataset with $n$ instances, such that $\mathcal{D} = \{(\vec{x}_{i}, y_i) ~|~ 1 \leq i \leq n\}$. Each instance $\vec{x}_{i} = \left[v_{i1},v_{i2},\ldots,v_{id}\right]$ is a vector with $d$ predictive attribute values and a target attribute, $y_i$. A meta-feature $f$ is a function $f: \mathcal{D} \to \mathds{R}^k$ that, when applied to a dataset $\mathcal{D}$, returns a set of $k$ values that characterize the dataset, and that are predictive for the performance of algorithms when they are applied to the dataset. Function $f$ can be detailed  as  \[f(\mathcal{D})~=~\sigma(m(\mathcal{D},~h_m),~h_s),\] such that $m: \mathcal{D} \to \mathds{R}^{k'}$ is a characterization measure; $\sigma: \mathds{R}^{k'} \to \mathds{R}^k$ is a summarization function; $h_m$ and $h_s$ are hyperparameters used for $m$ and $\sigma$, respectively. The summarization function is required in propositional scenarios when a fixed  cardinality is needed, given that $k$ is always constant regardless of the value of $k'$.

Traditionally, no distinction has been made between the concepts of a meta-feature, $f$, and a characterization measure, $m$.
This may be natural when a measure results in a single value ($k' = k = 1$) and $\sigma$ is the identity function, thus $f = m$. 
However, when a measure $m$ can extract more than one value from each dataset, i.e. $k'$ can vary according to $\mathcal{D}$, these values still need to be mapped to a vector of fixed length $k$. 
For instance, many authors use $f~\approx~\mathit{mean}(m)$ \citep{Ali2006,Castiello2005,Sohn1999}. 
Other common summarization functions are histograms \citep{Kalousis1999}, minimum and maximum \citep{Todorovski2000}, and skewness and kurtosis \citep{Reif2012b}.

These definitions allow the categorization of meta-features in a well-defined taxonomy, illustrated in Table \ref{tab:taxonomia}. 
In this framework, categories are divided into two groups, \emph{input} and \emph{output}, which are related to the characterization of the input and output of a measure, respectively. 
While some of these categories are only descriptive, others define whether or not a meta-feature is suitable for a specific scenario.

\begin{table}[t]%
\centering
\footnotesize
\begin{tabular}{lll}
\toprule
Level & Category Name & Options \\
\midrule
Input 
  & Task & 
  \begin{tabular}[t]{@{}l@{}}Classification \\ Supervised \\ 
  Any \end{tabular} \\ \addlinespace 
  & Extraction & 
 \begin{tabular}[t]{@{}l@{}} Direct \\ Indirect \end{tabular} \\  \addlinespace
  & Argument & 
  \begin{tabular}[t]{@{}l@{}}
   n Predictive Attribute (nP) \\ 
   All Predictive Attributes ($\ast$P)\\
   Target Attribute (T)
   \end{tabular} \\ \addlinespace 
  & Domain & 
  \begin{tabular}[t]{@{}l@{}} Numerical \\ Categorical \\ Both \end{tabular} \\ \addlinespace 

  & Hyperparameters & 
  \begin{tabular}[t]{@{}l@{}} Yes, No \end{tabular} \\   \addlinespace  \addlinespace

Output 
  & Range & [min, max] \\ \addlinespace
  & Cardinality & 
  $k$ \\ \addlinespace
  & Deterministic & 
  \begin{tabular}[t]{@{}l@{}} Yes, No \end{tabular} \\ \addlinespace
  & Exception & 
  \begin{tabular}[t]{@{}l@{}} Yes, No \end{tabular} \\ 

\bottomrule

\end{tabular}
\caption{Categories used to describe a measure or group of measures.}
\label{tab:taxonomia}
\end{table}

Some measures are restricted to specific tasks, such as \emph{classification}. 
Others can be more generically applied to \emph{supervised} tasks, such as regression problems. 
The measures classified as \emph{any} are the most general and can also be applied to unsupervised tasks such as clustering and semi-supervised problems. 
In \emph{supervised} and \emph{classification} tasks, a target attribute is required to evaluate the meta-features, which is not necessary for meta-features of the type \emph{any}. 

The cardinality defines the number of possible values in a measure. 
A distinction between single-valued measures ($k=1$) and multi-valued measures ($k>1$) is important for data analysis, mainly to define whether or not a summarization function must be applied. 
For most of the multi-valued measures, the cardinality is related to aspects such as instances, attributes or classes in the considered datasets.

Although most of the measures are \emph{deterministic}, some of them are \emph{non-deterministic}, thus there is no guarantee that the same result will be obtained for the same input in different runs. 
When reproducibility is necessary, the same seed must be used for each run or the measures must be executed a number of times to decrease the randomization effect.

Finally, while some measures are \emph{robust}, others can generate \emph{exceptions} for certain datasets, leading them not to emit valid values in all cases. 
This can occur in particular conditions, such as a division by zero or a logarithm of a negative number. 
The identification of these measures, the cases where they may not work as desired and alternatives to handle these situations are open issues on MtL.

\section{Meta-Features}
\label{sec:metafeatures}

A fundamental MtL question is: how to extract suitable information to characterize specific tasks? 
Researchers have been trying to answer this question by looking for dataset properties that can affect learning algorithm performance, measuring this performance outright~\citep{Bensusan2000,Pfahringer2000}, investigating alternatives~\citep{Kopf2000,Soares2001a} and adapting/creating new measures based on existing ones~\citep{Castiello2005,Sohn1999}. 

In all cases, the meta-features were always organized in groups. 
These groups are subsets of data characterization measures~\citep{Brazdil2009} that share similarities among them. 
However, the frontiers between them are not always clear and strictly delimited. 
The fact that two studies mentioned using the same group of measures does not mean that they used exactly the same measures~\citep{Smith-Miles2008}. 
Additionally, different names have been used to describe groups of the same measures.
In this work, we propose the organization of the measures in six groups:

\begin{description}

	\item[Simple:] measures that are easily extracted from data~\citep{Reif2014}, commonly known, and do not require significant computational resources~\citep{Reif2012}. They are also called \emph{general} measures~\citep{Castiello2005}.

	\item[Statistical:] measures that capture the statistical properties of the data \citep{Reif2014}. These measures capture data distribution indicators, such as average, standard deviation, correlation and kurtosis. They only characterize numerical attributes \citep{Castiello2005}.

	\item[Information-theoretic:] measures from the information theory field \citep{Castiello2005}. These measures are based on entropy \citep{Segrera2008}, capturing the amount of information in the data and their complexity \citep{Smith-Miles2008}. They can be used to characterize discrete attributes.

	\item[Model-based:] measures extracted from a model induced using the training data \citep{Reif2014}. They are often based on properties of decision tree (DT) models \citep{Bensusan2000,Peng2002}, when they are referred to as \emph{decision-tree-based} meta-features \citep{Bensusan2000}. Properties extracted from other models have also been used \citep{Filchenkov2015}.
    
	\item[Landmarking:] measures that use the performance of simple and fast learning algorithms to characterize datasets \citep{Smith-Miles2008}. The algorithms must have different biases and should capture relevant information with a low computational cost. Different approaches have been investigated \citep{Furnkranz2001,Soares2001a}.

    \item[Others:] measures not included in the previous groups, such as standalone measures, time-related measures~\citep{Reif2011}, concept and case-based measures~\citep{Munoz2018,Vanschoren2012}, clustering and distance based measures~\citep{Pimentel2019,Vukicevic2016}, among others. 
    Although they have been used in a small number of studies, they characterize aspects of the dataset that do not fit into the other groups.

\end{description}

The first three groups represent the most common and traditional approaches for data characterization \citep{Brazdil2009}. 
They receive different names such as \emph{basic} measures \citep{Filchenkov2015}, \emph{DCT} \citep{Peng2002}, \emph{standard} \citep{Engels1998} and \emph{STATLOG} measures \citep{Smith-Miles2008}. 
The next two require using machine learning algorithms because they extract model complexity or performance measures. 
The last group includes a wide diversity of measures, such as standalone measures, very expensive meta-features, measures suitable for particular scenarios, and small subgroups of measures.

\citeA{Lindner1999} describe a group, called \emph{discriminant} meta-features. However, in the meta-learning literature, the measures in this group are commonly called statistical meta-features.
\citeA{Vanschoren2010} uses a different categorization of meta-features, based on intrinsic biases of learning algorithms, such as \emph{data normality}, \emph{feature redundancy}, and \emph{feature-target association}.

In the remainder of this section, a systematic definition and description of these measures are provided, using the taxonomy shown in Table~\ref{tab:taxonomia}. 
The formal definition of each measure is available in \ref{sec:formalization}. 
In the descriptions, $-\infty$ and $\infty$ are used when it is not possible to define the range of a measure, whereas \emph{inherited} is used when the measure range is defined by the value range of specific dataset attributes. 
The use of an upper stroke bar in the range and cardinality indicates an approximated value. 
When the columns \emph{Extract}, \emph{Domain}, \emph{Hyperp.}, \emph{Excep.} and \emph{Det.} describe a constant property, they are suppressed from the tables and identified in the caption.
The section finishes with a description and an analysis of the main summarization functions.

\subsection{Simple meta-features}
\label{subsec:simple}

The simple measures are directly extracted from the data and they represent basic information about the dataset. They are the simplest set of measures in terms of definition and computational cost \citep{Castiello2005,Michie1994,Reif2012,Reif2014}. 
Table \ref{tab:simple} presents these measures. 
They are computed directly, free of hyperparameters and deterministic. 
Semantically, the measures represent concepts related to the number of predictive attributes, instances, target classes and missing values.

\begin{table}[!htp]%
\footnotesize
\setlength{\tabcolsep}{5pt}
\begin{minipage}{\textwidth}
\begin{center}
\scalebox{1}{
\begin{tabular}{lllllll}
  \toprule
  Acronym & Task & Argument & Domain & Range & Card. & Excep. \\
  \midrule

\emph{attrToInst} & Any & $\ast$P & Both & $[0, \overline{d}]$ & $1$ & No \\ 

\emph{catToNum} & Any & $\ast$P & Both & $[0, \overline{d}]$ & $1$ &  Yes \\

\emph{classToAttr} & Classif. & $\ast$P+T & Both & $[0, q]$ & $1$ &  No \\

\emph{freqClass} &  Classif. & T & Categ. & $[0,1]$ & $q$ & No  \\ 

\emph{instToAttr} & Any & $\ast$P & Both & $[0, \overline{n}]$ & $1$ & No  \\

\emph{instToClass} & Any & $\ast$P+T & Both & $[1, \overline{n}]$ & $1$ & No  \\

\emph{nrAttr} & Any & $\ast$P & Both & $[1, +\infty]$ & $1$ & No  \\

\emph{nrAttrMissing} & Any & $\ast$P & Both & $[0, d]$ & $1$ & No  \\

\emph{nrBin} & Any & $\ast$P & Both & $[0, d]$ & $1$ & No  \\

\emph{nrCat} & Any & $\ast$P & Both & $[0, d]$ & $1$ & No  \\

\emph{nrClass} & Classif. & T & Categ. & $[2, \overline{n}]$ 
& $1$ & No  \\

\emph{nrInst} & Any & $\ast$P & Both & $[q, +\infty]$  
& $1$ & No  \\

\emph{nrInstMissing} & Any & $\ast$P & Both & $[0, n]$ & $1$ & No  \\

\emph{nrMissing} & Any & $\ast$P & Both & $[0, \overline{dn}]$ & $1$ & No  \\

\emph{nrNum} & Any & $\ast$P & Both & $[0, d]$ & $1$ & No  \\

\emph{numToCat} & Any & $\ast$P & Both & $[0, \overline{d}]$ & $1$  & Yes  \\

  \bottomrule
\end{tabular}}
\end{center}
\end{minipage}
\caption{Simple measures and their characteristics. They are directly extracted, free of hyperparameter and deterministic.}
\label{tab:simple}
\end{table}%

The measures related to attributes are: number of attributes (\emph{nrAttr}); number of binary attributes (\emph{nrBin}); number of categorical attributes (\emph{nrCat}); number of numeric attributes (\emph{nrNum}); proportion of categorical versus numeric attributes (\emph{catToNum}) and vice-versa (\emph{numToCat}).
These measures are relevant to characterize the main aspects of a dataset, providing information that can support the choice of a data set for a particular learning task.

The number of instances (\emph{nrInst}) and the number of classes (\emph{nrClass}) by themselves do not provide much information since they indicate the dataset size and its label diversity.
However, when combined with the \emph{nrAttr}, different simple concepts can be captured.
The measures \emph{attrToInst} and \emph{instToAttr} represent the dimensionality and sparsity of the data, respectively.
The latter is a potential indicator for overfitting when its value is too small, a learning model may take into account irrelevant details in  the training data, resulting in poor generalization~\citep{Kuba2002};
the number of classes per attribute (\emph{classToAttr}) and instances per classes (\emph{instToClass}) relate properties of the dataset to the target attribute distribution.

Still concerning the target attribute, the frequency of instances in each class (\emph{freqClass}) allows the extraction of relevant measures, e.g. frequency of proportion of majority and minority class \citep{Ali2006}, default accuracy/error \citep{Peng2002} and standard deviation of the class distribution \citep{Lindner1999}.
When combined with summarization functions, it can describe imbalanced learning scenarios.

Finally, some measures assess dataset quality, such as the number of missing values in the dataset attributes (\emph{nrAttrMissing}) and instances (\emph{nrInstMissing}), as well as the total number (\emph{nrMissing}).
Since some ML algorithms can deal with missing values, these measures can provide important information for algorithm selection.
It must be observed that some measures cannot deal with missing values, and some previous data treatment is necessary, as discussed in Section~\ref{subsec:datatype}.

Some authors have proposed modified versions of these measures. 
As examples, \citeA{Todorovski2000} use the log of number of examples; \citeA{Brazdil1994} use the proportion of categorical attributes; \citeA{Kalousis2001a} use the proportion of numerical attributes. 
For a better meta-learning, it may be necessary to normalize the values of these measures. 
In these situations, with the exception of \emph{nrAttr} and \emph{nrInst}, the measures can be normalized using the theoretical maximum value, presented in the column Range.

It must also be observed that as some datasets have only categorical or numerical attributes, the \emph{catToNum} and \emph{numToCat} measures cannot be computed for all datasets.

\subsection{Statistical meta-features}
\label{subsec:statistic}

Statistical measures can extract information about the performance of statistical algorithms~\citep{Michie1994} or about data distribution, for instance, central tendency and dispersion~\citep{Castiello2005}. 
They are the largest and the most diversified group of meta-features, as shown in Table~\ref{tab:statistical}. 
Statistical measures are deterministic and support only numerical attributes. 
Some measures require the definition of hyperparameter values, while others can generate exceptions, e.g. caused by division by zero. 
Some of them are indirectly extracted, and are closely related to the discriminant group reported in~\citeA{Lindner1999}. 
The others can be widely applied since they use only predictive attributes as input.

\begin{table}[!t]%
\footnotesize
\setlength{\tabcolsep}{5pt}
\begin{minipage}{\textwidth}
\begin{center}
\scalebox{1}{
\begin{tabular}{lllllllll}
  \toprule
  Acronym & Task & Extract & Argument & Hyperp. &  Range & Card. & Excep. \\
  \midrule

\emph{canCor} & Classif. & Indirect & $\ast$P+T & No & $[0, 1]$ & $\overline{d}$ & No \\ 

\emph{cor} & Any & Direct & 2P & Yes & $[0, 1]$ & 
$\overline{d^2}$ & Yes  \\ 

\emph{cov} & Any & Direct & 2P & No & $[0, \infty]$ & 
$\overline{d^2}$ & No  \\ 

\emph{nrDisc} & Classif. & Indirect & $\ast$P+T & No & $[0, d]$ & $1$ & No \\

\emph{eigenvalues} & Any & Indirect & $\ast$P & No & $[0, \infty]$ & $\overline{d}$ & No \\

\emph{gMean} & Any & Direct & 1P & No & $[0, \infty]$ & $d$ & Yes  \\

\emph{hMean} & Any & Direct & 1P & No & \emph{inherited} & $d$ & No  \\

\emph{iqRange} & Any & Direct & 1P & No & $[0,\infty]$ & $d$ & No  \\

\emph{kurtosis} & Any & Direct & 1P & No & $[-3,\infty]$ & $d$ & Yes  \\

\emph{mad} & Any & Direct & 1P &  No &  $[0,\infty]$, & $d$ & No  \\

\emph{max} & Any & Direct & 1P & No & \emph{inherited} & $d$ & No  \\

\emph{mean} & Any & Direct & 1P & No & \emph{inherited} & $d$ & No \\

\emph{median} & Any & Direct & 1P & No & \emph{inherited} & $d$ & No  \\

\emph{min} & Any & Direct & 1P & No & \emph{inherited} & $d$ & No  \\

\emph{nrCorAttr} & Any & Direct & $\ast$P & Yes & $[0,1]$ & $1$ & Yes  \\

\emph{nrNorm} & Any & Direct & $\ast$P & Yes & $[0,d]$ & $1$ & No  \\ 

\emph{nrOutliers} & Any & Direct & $\ast$P & Yes & $[0,d]$ & $1$ & No \\ 

\emph{range} & Any & Direct & 1P & No & $[0,\infty]$ & $d$ & No  \\

\emph{sd} & Any & Direct & 1P &  No & $[0,\infty]$ & $d$ & No  \\

\emph{sdRatio} & Classif. & Indirect & $\ast$P+T & No & $[1, \infty]$ & $1$ & Yes  \\

\emph{skewness} & Any & Direct & 1P & No & $[-\infty,\infty]$ & $d$ & Yes  \\ 

\emph{tMean} & Any & Direct & 1P & Yes & \emph{inherited} & $d$ & No \\

\emph{var} & Any & Direct & 1P & No & $[0,\infty]$ & $d$ & No  \\

\emph{wLambda} & Classif. & Indirect & $\ast$P+T & No & $[0, 1]$ & $1$ & No  \\

  \bottomrule
\end{tabular}}
\end{center}
\end{minipage}
\caption{Statistical measures and their characteristics. They are deterministic and only accept numerical attributes.}
\label{tab:statistical}
\end{table}%

Correlation (\emph{cor}) and covariance (\emph{cov}) capture the interdependence of the predictive attributes~\citep{Michie1994}. 
They are computed for each pair of attributes in the dataset, resulting in $(d-1)/2$ values. 
The former is a normalized version of the latter, and the absolute value of both measures are frequently used, which changes the range from $[-1, 1]$ and $[-\infty, \infty]$, respectively, to the values reported in Table~\ref{tab:statistical}. 
High values indicate a strong correlation between the attributes, which can be interpreted as a level of redundancy in the data~\citep{Kalousis2001a}. 
To represent this information, \emph{nrCorAttr} computes the proportion of highly correlated attribute pairs.

Most statistical measures are extracted for each attribute separately. 
Measures of central tendency are composed by the \emph{mean} and its variations such as the geometric mean (\emph{gMean}), harmonic mean (\emph{hMean}) and trimmed mean (\emph{tMean}); and the \emph{median}. 
Measures of dispersion consist of the interquartile range (\emph{iqRange}), \emph{kurtosis}, maximum (\emph{max}), median absolute deviation (\emph{mad}), minimum (\emph{min}), \emph{range}, standard deviation (\emph{sd}), \emph{skewness} and variance (\emph{var}). 
While one points to the center of a distribution, the other shows how much the values are spread from the center, complementing themselves. 
Their range depends directly on the attributes' range, with few exceptions like kurtosis and skewness. 
These two, are suitable to capture the normality of the data attributes~\citep{Vanschoren2010}.

A specific measure to capture the normality of the attributes is the \emph{nrNorm}, which computes the number of attributes normally distributed.
Similarly, \emph{nrOutliers} counts the number of attributes that contain outliers.
Normality and outliers may impact the behavior of learning algorithms, which make these measures useful in an MtL scenario.

The discriminant statistical measures present some specificities such as being exclusively used for classification tasks. 
By considering the target value and using the whole dataset as input, they result in a single value. 
Canonical correlations (\emph{canCor}), the number of discriminant values (\emph{nrDisc}), the homogeneity of covariances (\emph{sdRatio}) and the Wilks lambda (\emph{wLambda}) represent the discriminant measures. 
Finally, the \emph{eigenvalues} from the covariance matrix only use the predictive data to be computed.

Concerning the hyperparameters, different correlation methods such as Pearson's correlation, Kendall's $\tau$ and Spearman's $\rho$ coefficient \citep{Rodgers1988}, can be used to compute the \emph{cor} measure. 
This is also applied to the \emph{nrCorAttr} measure, which additionally requires a threshold value to define high correlations. 
The \emph{tMean} requires the definition of how much data should be discarded to compute the mean. 
Finally, the \emph{nrNorm} and \emph{nrOutliers} are dependent on the algorithm to compute whether or not a distribution is normal and has outliers. 
Even though \emph{skewness} and \emph{kurtosis} could be seen as algorithm dependent, their variations do not produce observable differences for large samples of data \citep{Joanes1998}.

Some measures can throw exceptions and due to this are not calculated correctly. 
The \emph{cor}, \emph{kurtosis}, \emph{nrCorAttr} and \emph{skewness} could generate an error with a constant attribute caused by division by zero. 
The \emph{sdRatio} uses \emph{log} in this formulation, and the possibility of obtaining a negative value makes the measure error-prone. 
The \emph{gMean} can be computed in 2 different ways and both can generate errors, one using product and another using \emph{log}. 
The former can obtain arithmetic overflow/underflow while the latter cannot support negative values.

As the majority of the statistical measures do not consider the class information, \citeA{Castiello2005} proposed an indirect way to explore it. 
This approach splits the dataset according to the class labels and computes the measures for each subset. 
However, the authors are not aware of any empirical evaluation of this approach. 
Besides, many statistical measures need to be summarized since several possible values can be obtained.
Finally, it is important to observe that the statistical measures only support numerical attributes. 
Datasets that contain categorical data must be either partially ignored or converted to numerical values.

\subsection{Information-Theoretic meta-features}

Information-theoretic meta-features capture the amount of information in the data. 
Table~\ref{tab:infotheo} shows the information-theoretic measures, which require categorical attributes and most of them are restricted to representing classification problems. 
Moreover, they are directly computed, free of hyperparameter, deterministic and robust. 
Semantically, they describe the variability and redundancy of the predictive attributes to represent the classes.

\begin{table}[!htp]%
\footnotesize
\setlength{\tabcolsep}{5pt}
\centering
\begin{minipage}{\textwidth}
\begin{center}
\scalebox{1}{
\begin{tabular}{llllllll}
  \toprule
  Acronym & Task & Argument & Range & Card. \\
  \midrule

\emph{attrEnt} & Any & 1P & $[0, \log_2(n)]$ & $d$  \\

\emph{classEnt} & Classif. & T & $[0, \log_2(q)]$ & 1  \\

\emph{eqNumAttr} & Classif. & $\ast$P+T  & $[0,\infty]$ & 1  \\

\emph{jointEnt} & Classif. & 1P+T & $[0, \log_2(n)]$ & $d$  \\

\emph{mutInf} & Classif. & 1P+T  & $[0, \log_2(n)]$ & $d$  \\

\emph{nsRatio} & Classif. & $\ast$P+T & $[0,\infty]$ & 1  \\

  \bottomrule
\end{tabular}}
\end{center}
\end{minipage}
\caption{Information-theoretic meta-features and their characteristics. They are directly extracted, free of hyperparameter, robust, deterministic and support only categorical attributes.}
\label{tab:infotheo}
\end{table}%

The entropy of the predictive attributes (\emph{attrEnt}) and the target values (\emph{classEnt}) capture the average uncertainty present in the predictive and class attributes~\citep{Segrera2008}, respectively.
In the former, all predictive attributes are assessed, thus its summarization can provide an overview of the attributes' capacity for class discrimination.
In the latter, it represents how much information, on average, is necessary to specify one class~\citep{Castiello2005}.
In a learning perspective, a predictive attribute with a low entropy contains a low discriminatory power~\citep{Michie1994}, whereas a target attribute with low entropy contains a high level of purity.
These measures are usually normalized.

The joint entropy (\emph{jointEnt}) and the mutual information (\emph{mutInf}) compute the relationship of each attribute with the target values. 
While the former captures the relative importance of the predictive attributes to represent the target~\citep{Engels1998}, the latter represents the common information shared between them, indicating their degree of dependency~\citep{Michie1994}.

Finally, the equivalent number of attributes (\emph{eqNumAttr}) and the noise signal ratio (\emph{nsRatio}) capture information that is related to the minimum number of attributes necessary to represent the target attribute and the proportion of data that are irrelevant to describe the problem~\citep{Smith2001}, respectively.

To extract these measures from numerical attributes, we must know their data distribution or discretize them~\citep{Castiello2005}. 
The latter is simpler. However, being user-defined needs the introduction of hyperparameters, which is discussed further in Section \ref{subsec:datatype}.

\subsection{Model-Based meta-features}
\label{subsec:model-based}

The meta-features of this group are information extracted from a predictive learning model, in particular, a DT model. 
They characterize a dataset by how complex is the model induced, which, for DT, can be the number of leaves, the number of nodes and the shape of the tree.
Table \ref{tab:dtmodel} shows the DT model meta-features. 
They are designed to characterize supervised problems, all measures are deterministic, robust and require the definition of hyperparameters: the DT induction algorithm (together with its hyperparameter values) used to induce the DT model.

\begin{table}[!htp]%
\footnotesize
\setlength{\tabcolsep}{5pt}
\begin{minipage}{\textwidth}
\begin{center}
\scalebox{1}{
\begin{tabular}{lllll}
  \toprule
  Acronym & Task & Argument & Range & Card. \\
  \midrule

\emph{leaves} & Sup. & $\ast$P+T & $[q, \overline{n}]$ & 1 \\

\emph{leavesBranch} & Sup. & $\ast$P+T & $[1, \overline{n}]$ & $\overline{n}$  \\

\emph{leavesCorrob} & Sup. & $\ast$P+T & $[0, 1]$ & $\overline{n}$ \\

\emph{leavesHomo} & Sup. & $\ast$P+T & $[q, +\infty]$ & $\overline{n}$ \\

\emph{leavesPerClass} & Classif. & $\ast$P+T & $[0,1]$ & $q$ \\

s\emph{nodes} & Sup. & $\ast$P+T & $[q, \overline{n}]$ & 1 \\

\emph{nodesPerAttr} & Sup. & $\ast$P+T & $[0, \overline{n}]$ & 1 \\

\emph{nodesPerInst} & Sup. & $\ast$P+T  & $[0, 1]$ & 1 \\

\emph{nodesPerLevel} & Sup. & $\ast$P+T & $[1,\overline{n}]$ & $\overline{n}$  \\

\emph{nodesRepeated} & Sup. & $\ast$P+T & $[0, \overline{n}]$ & $\overline{d}$  \\

\emph{treeDepth} & Sup. & $\ast$P+T & $[1, \overline{n}]$ & $\overline{n}$  \\

\emph{treeImbalance} & Sup. & $\ast$P+T & $[0, 1]$ & $\overline{n}$  \\

\emph{treeShape} & Sup. & $\ast$P+T & $[0.0, 0.5]$ & $\overline{n}$  \\

\emph{varImportance} & Sup. & $\ast$P+T & $[0, 1]$ & $\overline{d}$ \\ 
  \bottomrule
\end{tabular}}
\end{center}
\end{minipage}
\caption{Model-based meta-features and their characteristics. These meta-features are indirectly extracted, robust, deterministic, require the definition of hyperparameters and support both attribute types.}
\label{tab:dtmodel}
\end{table}%

The measures based on leaves are identified with the prefix \emph{leaves}, which describe, in some degree, the complexity of the orthogonal decision surface. 
Some measures result in a value for each leaf, and those measures are the number of distinct paths (\emph{leavesBranch}), the support described in the proportion of training instances to the leaf (\emph{leavesCorrob}) and the distribution of the leaves in the tree (\emph{leavesHomo}).

The proportion of leaves to the classes (\emph{leavesPerClass}) represents the classes complexity and the result is summarized per class. While \emph{leavesCorrob} and \emph{leavesPerClass} have a fixed range independent of the dataset, \emph{leaves} and \emph{leavesBranch} have a maximum value limited by the number of instances. 
In practice, the most observed limit is associated with the number of attributes, which also determines the cardinality of them. 
Only \emph{leavesHomo} does not have a defined limit of values.

The measures based on nodes, which extract information about the balance of the tree to describe the discriminatory power of attributes, are identified with the prefix \emph{nodes}. 
Together with \emph{nodes}, the proportion of nodes per attribute (\emph{nodesPerAttr}) and the proportion of nodes per instance (\emph{nodesPerInst}) result in a single value. 
The number of nodes per level (\emph{nodesPerLevel}) and the number of repeated nodes (\emph{nodesRepeated}) have the number of attributes at their maximum value. 
While \emph{nodesPerLevel} describes how many nodes are present in each level, \emph{nodesRepeated} represents the number of nodes associated with each attribute used for the model.

The measures based on the tree size, which extract information about the leaves and nodes to describe the data complexity, are identified with the prefix \emph{tree}. 
The tree depth (\emph{treeDepth}) represents the depth of each node and leaf, the tree imbalance (\emph{treeImbalance}) describes the degree of imbalance in the tree and the shape of the tree (\emph{treeShape}) represents the entropy of the probabilities to randomly reach a specific leaf in a tree from each one of the nodes.

Finally, the importance of each attribute (\emph{varImportance}) represents the amount of information present in the attributes before a node split operation. 
The amount of information is defined by the randomization of incorrect labeling. 
This measure varies according to the DT algorithm.
As an example, the C4.5 algorithm uses the information gain from the information-theoretic group to compute the importance of the attributes \citep{Bensusan2000} and the CART algorithm uses the Gini index \citep{Loh2014}. 

Other model-based measures, using different learners, such as $k$-Nearest Neighbors (kNN) and Perceptron neural networks were presented in \citeA{Filchenkov2015}. 
However, some of these measures have a very high computational cost. 
Some others have the concept already described by well-known groups.
In~\citet{Nguyen2012}, the weights learned by distinct feature selection algorithms were defined as model-based meta-features.

\subsection{Landmarking meta-features}
\label{subsec:landmarking}

Landmarking is an approach that characterizes datasets using the performance of a set of fast and simple learners, different from the model-based meta-features, which extract information from the learning models.
Although the performance of any algorithm can be used as a landmarking, including sophisticated algorithms, some of them have been specifically used as meta-features. 
Table~\ref{tab:landmark} lists the most common landmarking measures. 
They characterize supervised problems and are indirectly extracted, thus the whole dataset is used as an argument. 
They require the definition of hyperparameters: the learning algorithm; the evaluation measure to asses the model performance; and, the procedure used to compute them (e.g. cross-validation). 
While the range is dependent on the evaluation measure (usually between 0 and 1), the cardinality is from the procedure, thereby it is user-defined. 
Since their training and test data samples are randomly chosen, all landmarking are non-deterministic.

\begin{table}[!htp]%
\footnotesize
\setlength{\tabcolsep}{5pt}
\begin{minipage}{\textwidth}
\begin{center}
\scalebox{1}{
\begin{tabular}{lllllll}
  \toprule
  Acronym & Task & Argument & Domain & Range & Card. & Excep.  \\
  \midrule

\emph{bestNode} & Sup. & $\ast$P+T & Both & $[0, 1]$  & \emph{user} & No  \\ 

\emph{eliteNN} & Sup. & $\ast$P+T & Both & $[0, 1]$ & \emph{user} & No  \\

\emph{linearDiscr} & Sup. & $\ast$P+T & Num. & $[0, 1]$ & \emph{user} & Yes  \\

\emph{naiveBayes} & Sup. & $\ast$P+T & Both & $[0, 1]$ & \emph{user} & No  \\

\emph{oneNN} & Sup. & $\ast$P+T & Both & $[0, 1]$ & \emph{user} & No  \\ 

\emph{randomNode} & Sup. & $\ast$P+T & Both & $[0, 1]$ & \emph{user} & No  \\

\emph{worstNode} & Sup. & $\ast$P+T & Both & $[0, 1]$ & \emph{user} & No  \\
  
 \bottomrule
\end{tabular}}
\end{center}
\end{minipage}
\caption{Common landmarking meta-features and their characteristics. They are indirectly extracted, non-deterministic and require the definition of hyperparameters.}
\label{tab:landmark}
\end{table}%

The measures \emph{bestNode}, \emph{randomNode} and \emph{worstNode} are the performance of a DT-model induced using different single attributes. 
Respectively, they use the following attributes: the most informative, a random one, and the least informative attribute. 
The aim is to capture information about the boundary of the classes and combine this information with the linearity of the DT-models induced with the worst and random attributes. 
The DT algorithm is a hyperparameter defined by the user since different algorithms could be used.

The elite-Nearest Neighbor (\emph{eliteNN}) is the result of the $1$NN model using a subset of the most informative attributes in the dataset, whereas the one-Nearest Neighbor (\emph{oneNN}) is the result of a similar learning model induced with all attributes. 
The distance measure used by the $k$NN algorithm is a hyperparameter.

The Linear Discriminant (\emph{linearDiscr}) and the Naive Bayes (\emph{naiveBayes}) algorithms use all attributes to induce the learning models. 
The first technique finds the best linear combination of predictive attributes able to maximize the separability between the classes. 
For such, it uses a covariance matrix and assumes that the data follow a Gaussian distribution. 
This technique can generate exceptions if the data has redundant attributes. 
The second technique is based on the Bayes' theorem and calculates, for each feature, the probability of an instance to belong to each class. 
The combination of all features and related probabilities for one instance returns the class with the highest probability.

Concerning the hyperparameters, an evaluation measure such as accuracy, balanced accuracy and Kappa is necessary to evaluate the models. 
Other measures such as precision, recall and F1 also could be used, however, for them, it is necessary to identify the class of interest in binary datasets. 
The procedures used to induce the model are \emph{(i)} using the whole instances to train and test; \emph{(ii)} holdout; and, \emph{(iii)} cross validation. 
This information is rarely mentioned in MtL studies and their impact in the characterization measures are not yet known. 
In practice, it represents a trade-off between stable measures and computational costs.

Some variants are relative and subsampling landmarkings~\citep{Furnkranz2001} and their combined use~\citep{Soares2001a}.
Instead of using the absolute performance of the landmarkers as meta-features, a relative approach adopts the landmarkers' ranking, which is obtained using pairwise comparisons. 
Thus, the meta-feature can be a binary value indicating the winner, the difference between them or the ratio of the two performances. 
Besides, a meta-feature for each ranking position containing the name of the respective landmarker is of categorical type.
On the other hand, subsampling landmarking works by applying traditional algorithms to a reduced subset of the original dataset.

The performance of the landmarkers can be represented as a learning curve, representing their use with different sampling sizes of a dataset~\citep{Leite2005}. 
Furthermore, in an algorithm recommendation scenario, their relative performance can be learned by meta-models and the prediction from these meta-models can be used as meta-features, analogous to a stacking-based approach~\citep{Sun2013}.

\subsection{Other meta-features} \label{misc}

Many other non-traditional characterization measures have been reported in the literature.
Despite the fact they are not broadly used in MtL studies, e.g. due to a high computational complexity or domain bias, they can be useful for a particular learning scenario and MtL problem. 
Besides, some works show good results when using those characterization measures \citep{Garcia2018, Morais2013, Pimentel2019}.
Here, they are arbitrarily presented in the following subgroups:  \emph{clustering and distance}; \emph{complexity}; and \emph{miscellaneous}.

\subsubsection{Clustering and distance-based}

Clustering and distance-based measures characterize the instance space using validation, also called index, measures that evaluate partitions produced by clustering algorithms and measures calculating the distance between instances.
Clustering validation measures and distance-based measures can be indirectly extracted characterization measures, requiring the set of hyperparameter values such as the clustering algorithm and the distance function, respectively.
With few exceptions, they are computed using only the predictive attributes.
According to the distance measure used, the meta-features can handle numerical and/or categorical attributes.
Table~\ref{tab:clustering} presents a list of clustering and distance-based measures.

\begin{table}[!htp]%
\footnotesize
\setlength{\tabcolsep}{5pt}
\begin{minipage}{\textwidth}
\begin{center}
\scalebox{1}{
\begin{tabular}{llllllll}
  \toprule
  Acronym & Task & Extract & Argument & Domain & Range & Card. &  Determ. \\
  \midrule

\emph{AIC} & Any & Indirect & $\ast$P & Both & $[0, \infty]$ & $1$ & No\\

\emph{BIC} & Any & Indirect & $\ast$P & Both & $[0, \infty]$ & $1$ & No \\

\emph{compactness} & Any & Indirect & $\ast$P & Both & $[0, \infty]$ & $\overline{n}$ & No \\ 

\emph{connectivity} & Any & Indirect & $\ast$P & Both & $[0, n]$ & $1$ & No \\ %

\emph{distInst} & Any & Direct & $\ast$P & Both & $[0, \infty]$ & $\overline{n^2}$ & Yes \\

\emph{distCorrInst} & Any & Direct & $\ast$P & Num. & $[0, 1]$ & $\overline{n^2}$ & Yes \\

\emph{gravity} & Classif. & Indirect & $\ast$P+T & Both & $[0, \infty]$ & $1$ & Yes \\

\emph{nrClusters} & Any & Indirect & $\ast$P & Both & $[1, \overline{n}]$ & $1$ & No \\

\emph{purityRatio} & Classif. & Indirect & $\ast$P+T & Both & $[0, 1]$ & $q$ & No \\
	
\emph{silhouette} & Any & Indirect & $\ast$P & Both & $[-1, 1]$ & $1$ & No \\ 

\emph{sizeDist} & Any & Indirect & $\ast$P & Both & $[0, 1]$ & $\overline{n}$ & No \\

\emph{XB} & Any & Indirect & $\ast$P & Both & $[0, \infty]$ & $1$ & No \\ 

  \bottomrule
\end{tabular}}
\end{center}
\end{minipage}
\caption{Clustering and distance-based characteristics. They are robust and require the definition of hyperparameters.}
\label{tab:clustering}
\end{table}%

Given the data partition produced by a clustering algorithm, \emph{nrCluster} represents the number of clusters, a simple informative measure, which is useful when this number is dynamically defined. 
When the clustering algorithm used has the number of clusters as a hyperparameter, a common option is to use the number of classes.
The distribution of the clusters based on the instances' frequency is captured by the measure \emph{sizeDist}.
A distribution skewed to the right indicates a complex dataset~\citep{Ler2018}.

Different validation measures are used to represent the quality of the partitions obtained, such as how compact each group is and how separated the groups are from each other~\citep{Vukicevic2016}. 
In a classification context, this information may indicate the separability of the instances, and possibly the classes. 
The Akaike Information Criterion (\emph{AIC}) and the Bayesian Information Criterion (\emph{BIC}) measures represent the relative quality of the partitions by estimating the amount of information lost by the model used to define the clusters.
For both, lower values indicate a better generalization of a model.
While \emph{Compactness} measures how compact the clusters are, \emph{Silhouette} and the Xie-Beni index (\emph{XB}) add separation between clusters to the compactness. 
The lower the value, the better for \emph{Compactness}, whereas, for \emph{Silhouette} and \emph{XB}, it is the opposite.
Other often-used clustering validation measures are presented next.

\emph{Connectivity} captures local densities by counting violations of the nearest neighbor  relationship of instances in different partitions~\citep{Handl2005}.
\new{When normalized by the number of instances, high values indicate that the clusters are not well separated.
It could be an informative measure to characterize the suitability of the bias related to instance-based learning algorithms.}

Although only these validation measures have been used to characterize datasets in MtL studies, there are many other clustering internal validation measures~\citep{Handl2005} that could be employed.
These measures can also be used without a clustering algorithm, by considering the classes as partitions.

Differently, \emph{purityRatio} is a clustering measure that looks at the instances' classes to evaluate the partitions.
It is calculated for each class and captures the ratio of clusters that contain instances related to the respective class.
Datasets with high values are more complex than those with low values since the classes are distributed across all partitions.

Another subset of measures, the distance-based measures~\citep{Pimentel2019} are obtained computing the distance between all pairs of instances (\emph{distInst}) and the correlations combined with the distances (\emph{distCorrInst}).
They indicate how close and related pairs of instances are, which may influence the decision boundaries of learning algorithms.
Finally, the center of gravity (\emph{gravity}) computes the dispersion among the groups of instances according to their class label. 
In this case, the groups are defined by the classes.

With few exceptions, all these measures have a high asymptotic computational complexity, which restricts their use. 
Additionally, they allow a wide number of choices, with different impacts in the value returned.
In spite of being able to provide a good characterization, clustering and distance measures are underexplored in the MtL literature.

\subsubsection{Complexity}

Complexity measures were proposed in~\citep{Ho2002} to capture the underlying difficulty of classification tasks, considering aspects such as class overlapping, the density of manifolds and the shape of decision boundaries. 
They were used to support data pre-processing, machine learning and recommender systems~\citep{Garcia2015, Garcia2018, Luengo2015, Smith2014}.
While the complete survey of the complexity measures can be found in~\citet{Lorena2018}, Table~\ref{tab:complexity} summarizes the main characteristics of these measures.

\begin{table}[!htp]%
\footnotesize
\setlength{\tabcolsep}{5pt}
\begin{minipage}{\textwidth}
\begin{center}
\scalebox{1}{
\begin{tabular}{lcccccccc}
  \toprule
  Acronym & Task & Extract & Argument & Domain & Hyperp. &  Range & Card. & Determ. \\
  \midrule

\emph{clsCoef} & Classif. & Indirect & $\ast$P+T & Num. & Yes & $[0, 1]$ & $1$ & Yes  \\
\emph{graphDensity} & Classif. & Indirect & $\ast$P+T & Num. & Yes & $[0, 1]$ & $1$ & Yes  \\ 
\emph{F1} & Classif. & Direct & 1P+T & Both & No & $[0, 1]$ & $1$ & Yes  \\
\emph{F1v} & Classif. &  Indirect & $\ast$P+T & Both & No & $[0, 1]$ & $1$ & Yes  \\
\emph{F2} & Classif. & Direct & 1P+T & Num. & No & $[0, 1]$ & $1$ & Yes  \\
\emph{F3} & Classif. & Direct & 1P+T & Num. & No & $[0, 1]$ & $1$ & Yes  \\
\emph{F4} & Classif. & Direct & $\ast$P+T & Num. & No & $[0, 1]$ & $1$ & Yes  \\
\emph{Hubs} & Classif. & Indirect & $\ast$P+T & Num. & Yes & $[0, 1]$ & $1$ & Yes  \\
\emph{LSC} & Classif. & Direct & $\ast$P+T & Num. & No & $[0, 1 - \frac{1}{n}]$ & $1$ & Yes \\
\emph{L1} & Classif. & Indirect & $\ast$P+T & Num. & No & $[0, 1]$ & $1$ & Yes \\
\emph{L2} & Classif. & Indirect & $\ast$P+T & Num. & No & $[0, 1]$ & $1$ & Yes \\
\emph{L3} & Classif. & Indirect & $\ast$P+T & Num. & No & $[0, 1]$ & $1$ & No  \\
\emph{N1} & Classif. & Indirect & $\ast$P+T & Num. & No & $[0, 1]$ & $1$ & Yes \\
\emph{N2} & Classif. & Direct & $\ast$P+T & Both & No & $[0, 1]$ & $1$ & Yes  \\
\emph{N3} & Classif. & Direct & $\ast$P+T & Both & No & $[0, 1]$ & $1$ & Yes  \\
\emph{N4} & Classif. & Direct & $\ast$P+T & Both & No & $[0, 1]$ & $1$ & Yes  \\
\emph{T1} & Classif. & Direct & $\ast$P+T & Num. & No & $[0, 1]$ & $1$ & Yes  \\
\emph{T2} & Any & Direct & $\ast$P & Both & No & $[0, \overline{n}]$ & $1$ & Yes  \\
\emph{T3} & Any & Indirect & $\ast$P & Num. & No & $[0, \overline{n}]$ & $1$ & Yes  \\
\emph{T4} & Any & Indirect & $\ast$P & Num. & No & $[0, 1]$ & $1$ & Yes  \\

  \bottomrule
\end{tabular}}
\end{center}
\end{minipage}
\caption{Complexity measures and their characteristics. They are robust measures.}
\label{tab:complexity}
\end{table}%

\citet{Ho2002} divide the complexity measures into three groups: (i) feature overlapping measures; (ii) measures of the separability of classes; and (iii) geometry, topology and density of manifolds measures. 
Following \citet{Lorena2018}, we adopted a more granular organization:
(i) feature-based measures; (ii) linearity measures; (iii) neighborhood measures; (iv) network measures; and (v) dimensionality measures. 

Feature overlapping measures characterize how informative the predictive attributes are to separate the classes. 
They are: maximum Fisher’s discriminant ratio (\emph{F1}); directional-vector maximum Fisher’s discriminant ratio (\emph{F1v}); volume of overlapping region (\emph{F2}); maximum individual feature efficiency (\emph{F3}); collective feature efficiency (\emph{F4}). 
The complexity is low if at least one predictive attribute can separate the classes.

Linearity measures quantify whether the classes are linearly separated.
They include sum of the error distance by linear programming (\emph{L1}); error rate of linear classifier (\emph{L2});  non-linearity of a linear classifier (\emph{L3}). 
To obtain the linear classifier, a linear Support Vector Machine (SVM) is often used.

Neighborhood measures analyze the neighborhoods of individual examples and try to capture class overlap and the shape of the decision boundary. 
They include fractions of Borderline Points (\emph{N1}); ratio of intra/extra class nearest neighbor distance~(\emph{N2}); error rate of the nearest neighbor classifier~(\emph{N3}); non-Linearity of the nearest neighbor classifier (N4); fraction of hyperspheres covering data~(\emph{T1}); local set average cardinality~(\emph{LSC}). 
All of them use a distance matrix between all pairs of points in the dataset to define the instances' neighborhoods according to their classes.

The network measures transform a dataset into a graph and extract structural and statistical information from the graph. 
In this new representation, each example from the dataset corresponds to a node, whilst undirected edges connect pairs of examples and are weighted by the distances between them. 
These measures include average density of the network~(\emph{graphDensity}) and Hub score~(\emph{hubs}). 
Other complex network measures are presented by~\citet{Morais2013}, however they are not detailed and we did not find other works using them.

Finally, the dimensionality measures evaluate data sparsity according to the number of instances relative to the predictive attributes of the dataset. 
The measures include the average number of points per dimension (\emph{T2}); the average number of points per PCA dimension (\emph{T3}); the ratio of the PCA dimension to the original dimension (\emph{T4}). 
While \emph{T2} is the \emph{instToAttr} meta-features, the \emph{T3} and \emph{T4} differ from \emph{T2} by using a transformed dataset instead of the original.

These complexity measures look at different complexity aspects in a dataset.
Thus, they can be related to other groups of measures presented in this study.
A variation of them to characterize the classes individually instead of the whole dataset is found in~\citet{Barella2018}. 
They are appropriate to represent the complexity of imbalanced datasets.
These complexity measures are free of hyperparameters and do not require the use of summarization functions, since some of them directly adopt a summarization procedure, e.g. \emph{F1} which uses the maximum value.
However, their extraction usually has a high computational cost, which restricts their use in MtL studies.

\subsubsection{Miscellaneous}

In this section, we included other characterization measures found in our review, which did not fit in the previous groups and were used in a small number of MtL studies. 
These measures are summarized in Table~\ref{tab:miscellaneous}.

\begin{table}[!htp]%
\footnotesize
\setlength{\tabcolsep}{5pt}
\begin{minipage}{\textwidth}
\begin{center}
\scalebox{.86}{
\begin{tabular}{lcccccccc}
  \toprule
  Acronym & Task & Extract & Argument & Domain & Hyperp. &  Range & Card. & Determ. \\
  \midrule

\multicolumn{9}{l}{\emph{Data distribution measures}} \\
\emph{attrConc} & Any & Direct & 2P & Categ. & No & $[0, 1]$ & $\overline{d^2}$ & Yes  \\

\emph{classConc} & Classif. & Direct & 1P+T & Categ. & No & $[0, 1]$ & d & Yes  \\

\emph{propPCA} & Any & Indirect & $\ast$P & Num. & Yes & $[0, 1]$ & 1 & Yes \\

\emph{sparsity} & Any & Direct & 1P & Both & No & $[0,1]$ & $d$ & Yes  \\ \addlinespace

\multicolumn{9}{l}{\emph{Case base measures}} \\
\emph{consistencyRatio} & Supervised & Direct & $\ast$P+T & Both & No & $[0, 1]$ & 1 & Yes \\
\emph{incoherenceRatio} & Any & Direct & $\ast$P & Both & Yes & $[0, 1]$ & 1 & Yes \\
\emph{uniquenessRatio} & Any & Direct & $\ast$P & Both & No & $[0, 1]$ & 1 & Yes \\ \addlinespace

\multicolumn{9}{l}{\emph{Concept based measures}} \\
\emph{cohesiveness} & Classif. & Direct & $\ast$P+T & Both & Yes & $[0, \overline{n}]$ & $n$ & Yes \\
\emph{wgDist} & Any & Direct & $\ast$P & Both & Yes & $[0, \infty]$ & $n$ & Yes \\

\addlinespace

\multicolumn{9}{l}{\emph{Structural Information}} \\
\emph{oneItemset} & Any & Indirect & $\ast$P & Both & No & $[0, 1]$ & $d$ & Yes \\
\emph{twoItemset} & Any & Indirect & $\ast$P & Both & No & $[0, 1]$ & $\overline{d^2}$ & Yes \\ \addlinespace

\multicolumn{9}{l}{\emph{Time based measures}} \\
\emph{infotheoTime} & Any & Indirect & $\ast$P & Categ. & No & $[0, \infty]$ & $1$ & No \\
\emph{landTime} & Supervised & Indirect & $\ast$P+T & Both & No & $[0, \infty]$ & $\overline{7}$ & No \\ 
\emph{modelTime} & Supervised & Indirect & $\ast$P+T & Both & No & $[0, \infty]$ & $1$ & No \\ 
\emph{statTime} & Any & Indirect & $\ast$P & Num. & No & $[0, \infty]$ & $1$ & No \\
\addlinespace

  \bottomrule
\end{tabular}}
\end{center}
\end{minipage}
\caption{
Other miscellaneous measures and their characteristics. They are robust measures.}
\label{tab:miscellaneous}
\end{table}%

Data distribution measures assess how the data is distributed in the predictive attribute space.
One of these measures is the concentration coefficient, also known as \emph{Goodman and Kruskal's $\tau$} \citep{Kalousis2001a}, which is applied to each pair of attributes (\emph{attrConc}) and to each attribute and the class (\emph{classConc}). 
In the former $d(d-1)$ values are obtained, since it is not symmetric, whereas in the latter, $d$ values are obtained, given that each attribute is associated with the class. 
Semantically, they represent the association strength between the attributes in each pair of attributes and between each predictive attribute and the target attribute.

Other related measures are the proportion of principal components that explain a specific (e.g. 95\%) variance of the dataset (\emph{propPCA}) and the \emph{sparsity}, which extracts the degree of discreetness in each attribute. 
\new{The former is another measure for capturing the redundancy of predictive attributes, whereas, the latter indicates the variance in the values of the attributes.}

Case base measures compare the instances with each other to identify properties that might make the learning process more difficult~\citep{Kopf2002}. 
Most of them are originally proposed as logical measures, however, instead of only capturing the occurrence (or not) of each property, we propose small changes to quantify each occurrence.
The \emph{consistencyRatio} quantifies the proportion of repeated instances with different targets, where zero is an ideal value.
The \emph{uniquenessRatio} is a generalization of \emph{consistencyRatio}, since it uses only the predictive attributes.
To measure how dissimilar the instances are in their attribute space, \emph{incoherenceRatio} computes the proportion of instances that do not overlap with any other instances in a predefined number of attributes.
Values close to 1 are preferred in a dataset since it shows that the instances are scattered through the input space.

The concept-based measures characterize the sparsity and the irregularity of the input-output distribution~\citep{Vilalta2002a}. 
An irregular distribution is observed when neighboring instances have distinct target values~\citep{Munoz2018}.
\new{The weighted distance (\emph{wgDist}) captures how dense or sparse the distribution of the instances is~\citep{Vilalta1999}. 
It could be defined as a distance-based measure.}
\emph{Cohesiveness} measures the density of the example distribution~\citep{Vanschoren2010}.
Another measure of this subgroup, the concept variation~\citep{Vilalta2002a} is defined by the cohesiveness average of all possible instances in the input space, therefore unfeasible.
Its version using the existing instances is captured by the summarization function \emph{mean}.

\new{Structural information works well in identifying similar datasets~\citep{Wang2015}, by characterizing binary itemsets to capture the distribution of values of both single attributes (\emph{oneItemset}) and pairs of attributes (\emph{twoItemset})~\citep{Song2012}.
They capture different and complementary aspects of the dataset. 
\emph{oneItemset} captures information of each individual's attributes, whereas, \emph{twoItemset} captures possible correlations concerning pairs of attributes.}
Association rules can also be applied to the transformed dataset to characterize other relations between attributes~\citep{Burton2014,Munoz2018}.

Time-based measures comprise the elapsed time to compute the previous groups of measures~\citep{Reif2011}, such as statistical, information-theoretic, model-based and landmarking.
In this case, the same hardware should be used to compute the meta-features from different datasets, which can be very restrictive.
Another option is to use the number of float point operations, but it is not always possible.

\subsection{Summarization Functions}
\label{sumfunc}

In this study, the purpose of summarization functions is to normalize the cardinality of meta-features and to characterize other meta-feature aspects, such as tendency, distribution and variability of the results. 
Given that many measures are multi-valued and that their cardinalities vary according to the dataset, comparisons between multiple datasets can be infeasible. 
Consequently, the summarization transforms non-propositional data to propositional \citep{Todorovski2000}, making them suitable to be organized in a meta-base, for instance. 
In the literature, summarization functions have been called meta-level attributes \citep{Todorovski2000}, meta$^2$-features \citep{Reif2012b} and post-processing functions \citep{Pinto2016}.

It is worth noting that in some studies~\citep{Castiello2005,Filchenkov2015,Kuba2002}, to cite a few, the mean function is used as part of the meta-feature definition and it is the only way used to summarize the results. 
Other studies have used distinct subsets of summarization functions, such as histogram~\citep{Kalousis1999}; minimum, mean and maximum~\citep{Todorovski2000}; minimum, maximum, mean and standard deviation~\citep{Feurer2014,Garcia2015,Peng2002a}; mean, standard deviation and quartiles 1, 2 and 3~\citep{Bilalli2018}; minimum, maximum, mean and standard deviation, kurtosis and skewness~\citep{Reif2012b}.

Table \ref{tab:summarization} presents a non-exhaustive list of the summarization functions, their range, cardinality and a brief description. 
The \emph{quantiles} and \emph{histogram} result in multiple values. 
The former summarizes a measure by representative values of the measure distribution, whereas the latter uses the proportion of values in each range of data. 
A hyperparameter specifying the number of bins in which the results are split~\citep{Kalousis1999} defines the cardinality of the \emph{histogram}. 
Some functions such as \emph{count}, \emph{histogram} and \emph{kurtosis} change the range of the characterized measure, while others inherit the range of the measure in which they summarize, such as \emph{max}, \emph{mean} and \emph{min}. 
The \emph{identity function} is conceptually used when a characterization measure results in a single value ($k'= 1$).

\begin{table}[!htp]%
\footnotesize
\begin{minipage}{\textwidth}
\begin{center}
\begin{tabularx}{\textwidth}{lccX}
  \toprule
  Acronym & Range & Cardinality & Brief description \\
  \midrule

\emph{count} & $[1,k]$ & 1 & Computes the cardinality of the measure, suitable when the cardinality is variable. \\
\emph{histogram} & $[0,1]$ & \emph{user} & Describes the distribution of the measured values, suitable for measures with high cardinality.\\
\emph{iqRange} & $[0,\infty]$ & 1 & Computes the interquartile range of the measured values.\\
\emph{kurtosis} & $[-3,\infty]$ & 1 & Describes the shape of the measured values distribution. \\ 
\emph{max} & \emph{inherited} & 1 & Results in the maximum values of the measure. \\
\emph{mean} & \emph{inherited} & 1 & Computes the averaged values of the measure. \\
\emph{median} & \emph{inherited} & 1 & Results in the central value of the measure. \\
\emph{min} & \emph{inherited} & 1 & Results in the minimum value of the measure. \\
\emph{quartiles} & \emph{inherited} & 5 & Results in the minimum, first quartile, median, third quartile and maximum of the measured values. \\
\emph{range} & $[0,\infty]$ & 1 & Computes the range of the measured values. \\
\emph{sd} & $[0,\infty]$ & 1 & Computes the standard deviation of the measured values. \\
\emph{skewness} & $[-\infty,\infty]$ & 1 & Describes the distribution shape of the measured values in terms of symmetry. \\

 \bottomrule
\end{tabularx}
\end{center}
\end{minipage}
\caption{
Main summarization functions.}
\label{tab:summarization}
\end{table}%

\citeA{Pinto2016}~proposed that the summarization functions should be organized in groups: \emph{descriptive statistical} includes the most common functions and summarizes a set of values in a single result such as \emph{max}, \emph{min}, \emph{mean}, \emph{median}, \emph{sd}, \emph{skewness}, \emph{kurtosis}, \emph{iqRange}, among others; \emph{distribution} characterizes the distribution of the measure using multiple values. For this purpose, the use of histogram with a fixed number of bins~\citep{Kalousis1999} and the use of quartiles to summarize the set of values~\citep{Bilalli2018} are alternatives observed in the literature; \emph{hypothesis test} assesses an assumption about a set of values, resulting in one or more values, as the p-values and/or the tests result. 
However, its use has not been observed in the literature.

Conceptually, any function that offers guarantees of a fixed cardinality, regardless of the number of values received by it, can be applied as a summarization function. 
Thus, even though a \emph{post-processing} function~\citep{Pinto2016} can also generate  indiscriminate number of values, a summarization function cannot. 
The summarization functions presented in Table~\ref{tab:summarization} can be applied to all multi-valued measures indiscriminately. 
Some combinations measure/summarization-function explore semantic concepts, e.g. the standard deviation of the classes proportion~\citep{Lindner1999}. 
Particular summarization functions, suitable for a specific measure, such as the \emph{nrCorAttr} statistical meta-feature, that summarizes the \emph{cor}, is better instantiated as a meta-feature. 
Section~\ref{subsec:summarization} addresses this matter as an open issue and shows possible insights concerning their use and exploration.

\section{Discussion}
\label{sec:discussion}

In machine learning, it is expected that all information necessary to reproduce empirical experiments, obtaining similar results, should be clearly reported. 
For MtL, the information's need to maintain the reproducibility is even greater, since this research topic also includes all the machine learning analysis plus the recommendation system which is based on the characterization of several datasets and the performance assessment from a set of algorithms over the datasets.
However, many details related to them are frequently ignored or subtly addressed in the literature.

This section focuses on six aspects of the characterization process, most of them strictly related to the  taxonomy proposed in Section \ref{sec:taxonomy}.
Frequently ignored details, the unspoken decisions taken by researchers, are reviewed, along with the enumeration of gaps that demand further analysis whether theoretical, empirical or both.

\subsection{Input Domain}
\label{subsec:datatype}

The input domain defines the data type supported by a meta-feature. 
For instance, statistical meta-features support only numerical data while information-theoretic meta-features support only categorical data. 
The alternatives adopted to handle non-supported data types have rarely been reported in the literature, as observed in~\citeA{Smith2001,Ali2006,Reif2014,Garcia2015}. 
Besides the fact that such choices affect the reproducibility of MtL experiments, their impact on the outcomes is unknown.

\begin{figure}[!htb]
\includegraphics[width=\textwidth, angle=0]{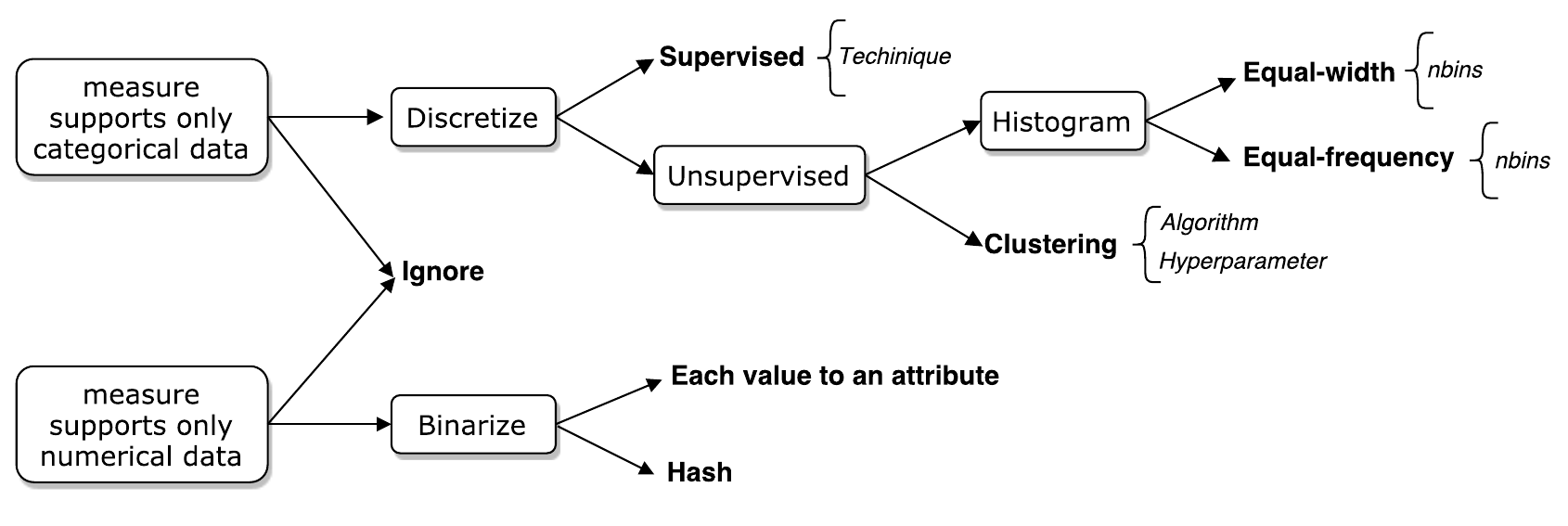}
\caption{Options to handle the input data type that are not supported by the meta-features.}
\label{fig:measure-domain}
\end{figure}

Figure \ref{fig:measure-domain} summarizes the options adopted in the literature to deal with the data type. 
The options consist of ignoring~\citep{Kalousis1999} or transforming the data~\citep{Castiello2005}. 
By ignoring the attributes, two problems are faced: \emph{(i)} if a dataset contains only attributes with the ignored data type, all respective measures will have missing values; \emph{(ii)} in an MtL context, the algorithms/techniques recommended may support the ignored data. 
In favor of this choice, it can be argued that to using only the meta-features that are able to characterize such data is a natural choice since they can properly represent the data~\citep{Michie1994}. 
Besides, their inability to process some types of data may be aligned with the limitations of some algorithms, therefore representing useful information.
Alternatively, the datasets can be segmented by type (only numerical, only categorical and mixed) where only the suitable measures for each group are used~\citep{Bilalli2017,Kopf2002}. 

By transforming the attributes, the meta-features can support any data types using a \emph{binarization} or \emph{discretization} approaches. 
It leads to new decisions since there are different alternatives used to transform the data, including the possibility of combining them together.

The most common transformation of categorical attributes into numerical ones is called \textit{binarization}~\citep{Aggarwal2015}. 
In this process, $\phi$ new binary attributes are created to represent each different category in the data, where $\phi$ is the number of distinct categories in the attribute. 
For each instance, only one of the new attributes is assigned to ``1'' while the others are assigned to ``0''. 
Its use to transform categorical attributes with a high number of distinct values is not recommended, since it generates a large number of new attributes. 
Alternatively, each category can be mapped to an integer and then represented in a binary hash, where $\log_2(\phi)$ new attributes are used to represent the bits values of the represented information~\citep{Tan2005}. 
The unintended relationships among the new attributes can be a deficiency of this approach, considering the meaninglessness of these relations.

Similarly, some meta-features support only categorical attributes, and the transformation from numeric to categorical attributes may be necessary. 
For such, \emph{discretization} techniques can be used. These techniques distribute numeric values in $\phi$ distinct intervals, which correspond to the new categories~\citep{Aggarwal2015}. 
As a result, order relations in the original values and variations within the same interval are lost.
In an unsupervised approach, the intervals can be defined using \emph{equal-width} or \emph{equal-frequency}, where they have the same interval width or the number of values, respectively. 
Other techniques such as clustering, correlation analysis and decision tree analysis can also be used for value discretization~\citep{Fayyad1993,Han2005}. 
The last two, which are supervised approaches, use the target attribute to define the categories.

The discretization procedure has a larger number of alternatives than the binarization procedure, which makes the result even more biased when they are arbitrary-defined. 
Most known methods are based on supervised and unsupervised techniques. 
The unsupervised techniques include the histogram and the clustering strategy.
Given that in each transformation there is a loss of information and a good discretization process can minimize it~\citep{Jin2007}. 
Because the unsupervised approaches are the simplest alternatives to discretize the data, more information are lost in the process, however, they have a lower cost than the supervised approaches.

The presence of missing values in the original datasets also demands attention, considering that many meta-features do not support the defective records. 
The alternatives to address this issue are: \emph{(i)} imputation of values provided by a preprocessing step and \emph{(ii)} removal of attributes and/or records with missing values. 
This topic is also frequently ignored in MtL papers.

\subsection{Hyperparameter values}
\label{subsec:userparam}

Another aspect that impacts the reproducibility of MtL experiments is the lack of details with regards to the hyperparameter values required by the measures. 
Possibly, this occurs because a value is used by default.

Tables \ref{tab:statistical}, \ref{tab:dtmodel}, \ref{tab:landmark} and \ref{tab:miscellaneous} identify the measures that require the definition of hyperparameter values. 
Some statistical measures have specific hyperparameter values.
All model-based and landmarking meta-features, on the other hand, have hyperparameter values that affect the whole group. 
For the model-based, different DT algorithms can be used to induce the model and each algorithm requires additional configurations. 
For the landmarking, the validation strategy, the evaluation measure and also the algorithms hyperparameters can be modified. 
In these cases, the same set of configurations is usually adopted for all measures of the group, but not necessarily by more than one author.

Other decisions concerning the use of meta-features and summarization functions can also be seen as hyperparameters. 
For instance, how to handle the unsupported data type,  as described in Subsection~\ref{subsec:datatype}, and the transformation by class~\citep{Castiello2005} proposed to explore the target information,  affect the statistical and information-theoretic groups and can also be defined as hyperparameters.
Additionally, the \emph{histogram} summarization function also has a hyperparameter that defines the number of bins to represent the measures.

In summary, the effects of such choices in the data characterization process are unknown. 
Alternatives, such as tuning the different parameters of the measures, using distinct instances of the same measure and evaluating the amount of information captured by them, have not been explored.


\subsection{Range of the Measures}
\label{subsec:datarange}

\begin{figure}[!htb]
\centering
\includegraphics[width=0.75\textwidth,angle=0]{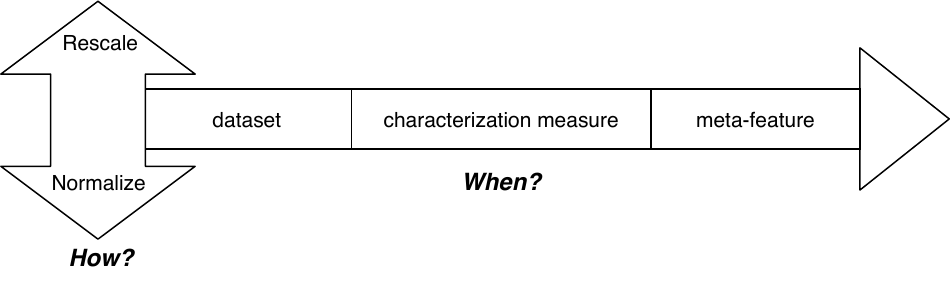}
\caption{Options to transform the range of the measures.}
\label{fig:measure-range}
\end{figure}

The data range has been frequently ignored in MtL studies, which suggests that meta-features have been used directly without transformation or it has not been properly reported. 
Although meta-features have a different range of values, they are used together in a meta-base. 
Considering that some algorithms are influenced by attributes with different ranges~\citep{Han2005,Wang2013}, the meta-data can be transformed by min-max scaling or z-score normalization, as illustrated by the vertical axis in Figure~\ref{fig:measure-range}.

The transformation can occur in three distinct moments: \emph{(i)} in the \emph{dataset}, before any computation; \emph{(ii)} in the result of the characterization measure, before the summarization function; and \emph{(iii)} in the meta-base, after computing the meta-feature. 
These moments are represented by the horizontal axis in Figure~\ref{fig:measure-range}. 
They have some implications in the result regardless of how the transformation occurs.

The dataset transformation is an alternative for the measures whose scale is determined by the values present in the dataset (range is \emph{inherit}). 
Changes in the original data range will reflect on the outcome of these meta-features. 
The second alternative transforms the result of the characterization measures. 
It is more suitable for multi-valued measures. 
Both alternatives are not recommended for meta-features using summarization functions on a particular scale, such as \emph{kurtosis} and \emph{skewness}.
Finally, the most conventional approach is to transform the meta-features result, which requires the characterization of all datasets beforehand.

Some rescaled meta-features are used along with (or instead of) their original version. 
The proportion of numeric and categorical attributes~\citep{Brazdil1994,Kalousis2001a}, the proportion of attributes with outliers and normal distribution~\citep{Brazdil2003,Salama2013} and the normalized entropy~\citep{Castiello2005}, are some examples found in the literature. 
However, only a few measures have their rescaled version named. 
The theoretical maximum and minimum values from the measures with a non-infinity range can be modified with the min-max scaling. 
The transformation of meta-features for some dataset characteristic (e.g. the number of instances) using absolute or relative values can be a better alternative.

In summary, the lack of information about the procedures adopted concerning the meta-data transformation is also a barrier to reproducible MtL studies. 
The different alternatives to transform the meta-features can suit some meta-features better than others. 
Although this topic does not contribute directly to the reproducibility issue, it is a very important research question that has not been satisfactorily addressed in the MtL literature.

\subsection{Summarization Functions}
\label{subsec:summarization}

In most MtL studies, summarization functions are combined with meta-features, either implicitly or explicitly. 
Implicitly when they are defined as part of the meta-feature formalization~\citep{Castiello2005,Filchenkov2015,Kuba2002,Peng2002a}, where the average result is the most natural solution used. 
Explicitly when studies show the effectiveness of using other options to summarize measures~\citep{Kalousis1999,Pinto2016,Reif2012b,Todorovski2000}, as reported in Section~\ref{sumfunc}.

Some combinations of meta-features and summarization functions have a semantic meaning. 
For instance, the  standard deviation (\emph{sd}) applied to the frequencies of the classes (\emph{freqClass}) shows how uniform the class distribution is, which may also indicate that the classes are unbalanced.
Other combinations are meaningless, as the use of the cardinality of the measure (\emph{count}) to summarize the joint entropy (\emph{jointEnt}), since the measure has a fixed cardinality. 
There are also some possible problematic combinations, such as the use of \emph{histograms} to summarize meta-features with low cardinality and/or with the range that is defined according to a dataset characteristic. In this case, the histogram bins can be sparse and represent different scales of values for each dataset.

The use of many functions to summarize a measure proportionally increases the number of meta-features obtained. 
As many measures are multi-valued, hundreds of results can be easily obtained when combined with multiple summarization functions. 
The relatively low number of meta-instances usually observed in MtL experiments together with the high number of meta-features could generate meaningless models due to the curse of dimensionality~\citep{Tan2005}. 
The use of a feature-selection algorithm can be an alternative to deal with this problem~\citep{Lemke2015,Pinto2016}.

Even though summarization functions are not strictly related to reproducibility issues, they are relevant to reproducibility because different choices can be made in a characterization process. 
The empirical analysis of summarization functions and the exploration of new ways to summarize meta-features should be the subject of future research.

\subsection{Exceptions}
\label{subsec:anomalies}

As discussed previously, some measures can be incorrectly computed for some datasets. 
Their use requires specific conditions that cannot always be guaranteed. 
Operations such as division by zero and logarithm of negative values are the main causes of exceptions. 

Alternatives to deal with problematic measures are: \emph{(i)} assuming it results in a missing value; \emph{(ii)} using a default value; \emph{(iii)} if the measure is multi-valued, ignore it. 
The first option results in a meta-base with missing values, which eventually will be filled using some preprocessing technique~\citep{Han2005} or removed from the meta-base. 
The other two alternatives fix the problem of having a missing value during the computation of the meta-feature.

The use of a default value to represent exceptional cases can be positive when it properly characterizes the measure and the phenomenon that generates the exception. 
Table~\ref{tab:error-prone1} presents default values, suggested by the authors, to be used when a meta-feature cannot characterize a dataset. 
With the exception of~\emph{sdRatio}, the values are in the range of their measures, assuming a semantic meaning as explained in the column \emph{Meaning}.

\begin{table}[!htp]%
\begin{minipage}{\textwidth}
\begin{center}
\begin{tabular}{llcl}
\toprule
 Group & Measure & Default & Meaning \\
\midrule

\multicolumn{4}{l}{\emph{Mono-valued measures}} \\
Simple & \emph{catToNum} & \texttt{d} & All attributes are categoric. \\
 & \emph{numToCat} & \texttt{d} & All attributes are numeric. \\\addlinespace
 Statistical & \emph{nrCorAttr} & \texttt{0} & No pair of attributes is highly correlated. \\
 & \emph{sdRatio} &\texttt{-1} & Invalid result. \\

\midrule

\multicolumn{4}{l}{\emph{Multi-valued measures}} \\
 Statistical & \emph{cor} & \texttt{0} & No correlation. \\
 & \emph{gMean} & \emph{mean} & Mean value. \\
 & \emph{kurtosis} & \texttt{0} & Constant values. \\
 & \emph{skewness} & \texttt{0} & Constant values. \\\addlinespace
 Landmarking & \emph{linearDiscr} & \texttt{0} & Low predictive performance. \\ 
 \bottomrule
\end{tabular}
\end{center}
\end{minipage}
\footnotesize
\caption{Suggested values to fill the missing cases for the meta-features with exceptions.}
\label{tab:error-prone1}
\end{table}%

The previous alternatives can introduce noise in the predictive meta-data. 
This does not occur when the defective results can be removed before the summarization. 
As a drawback, this alternative is valid only for the multi-valued measures. 
Furthermore, to discard few values for measures with high cardinality, the final result will not change drastically, but for the measures with low cardinality, this approach may lead to distortions in results.

Summarization functions can also generate exceptions. 
This is the case of \emph{sd}, \emph{kurtosis} and \emph{skewness}. 
The \emph{sd} cannot be applied to single values while the \emph{kurtosis} and \emph{skewness} cannot be applied to constant vectors.
The alternatives \emph{i} and \emph{ii} can also be adopted for them. 
The value 0 is the default value suggested to fill the problematic cases, which represents no deviations for \emph{sd} and constant values for \emph{kurtosis} and \emph{skewness}.

In summary, the use of these measures and summarization function does not imply that they will generate exceptions during the extraction of meta-features.
However, there is an absence of information about the occurrence or lack of occurrence in empirical studies in MtL.
Thereby, it is strictly related to the reproducible of the MtL studies, given that it has a technical bias and is related to the implementation and use of meta-features.

\subsection{Meta-feature Space}
The ratio between the number of meta-features and the number of meta-instances in MtL experiments is usually higher than in conventional ML experiments.
Furthermore, it is well known that the most suitable meta-features varies for different MtL tasks~\citep{Bilalli2017}.
Thus, some studies have investigated the use of feature selection techniques~\citep{Pinto2016,Salama2013} and the transformation of the meta-features' space~\citep{Bilalli2017,Munoz2018} to reduce the dimensionality of meta-bases, as well as to increase the predictive performance of meta-models~\citep{Kalousis2001}.

Meta-feature selection is just an instance of feature selection~\citep{Lemke2015}.
Among the different approaches for meta-feature selection, wrapper appeared more often in our literature review~\citep{Todorovski2000,Kalousis2001,Brazdil2009,Reif2014,Filchenkov2015, Garcia2015} than the use of a filter~\citep{Peng2002, Lee2008, Pinto2016}.

In~\citet{Munoz2018}, the authors followed a new approach for meta-feature selection. They investigated the behaviour of several meta-features in $12$ classification challenges. 
By modifying a dataset to increase/decrease each investigated problem, the variance of the meta-features is statistically assessed, revealing those that better characterize each variation.
After the repetition of the process using different datasets, the most relevant features for each challenge are obtained.

Another work for the meta-feature dimensionality reduction used Principal Component Analysis (PCA)~\citep{Hotelling1933} to obtain latent meta-features~\citep{Bilalli2017}. 
After computing the principal components, the most relevant (according to the cumulative total variance) are selected.
The authors later used a filter based on correlation with the target to select a subset of the latent meta-features.

As PCA does not take into account the target variable to transform the data, \citet{Munoz2018} used optimization to transform a set of previously selected meta-features into a 2-D space. 
For such, the authors used the performance of several learning algorithms.
Named instance space, it enables the visualization of the set of datasets used in an MtL study.

Most of the studies found for this study use wrapper. 
Few studies use transformation approaches in MtL.
Some works have compared groups of meta-features~\citep{Abdelmessih2010,Kopf2002,Reif2011,Reif2014}, with different findings.
For instance, landmarkings and model-based meta-features were the most important characterization measures in~\citeA{Reif2014} and \citeA{Filchenkov2015}, respectively. 
In contrast, feature selection wrapper did not improve the predictive performance of the meta-models in~\citeA{Garcia2015}.

To estimate the importance of a meta-feature, \citeA{Filchenkov2015} uses a significance measure that associates the predictive performance of a model induced using each meta-feature alone. 
This process is repeated several times and the average performance obtained for each meta-feature is the meta-feature significance value.
\citeA{Pimentel2019} define the meta-feature importance as the number of times it is selected when the Random Forest algorithm is applied to the meta-base.
In \citeA{Salama2013,Peng2002}, the authors use the correlation between them and the meta-target to select the meta-features~\citep{Peng2002,Salama2013}.

The decision of whether to use reduction and/or transformation is an important issue in the reproducibility and performance of MtL experiments.
When used, a detailed specification of the procedures adopted is essential for the replication of the experiments.
Moreover, while meta-feature selection may improve the interpretability of the meta-models, the same is not the case when a transformation is used.

\subsection{Outline}
\label{subsec:outline}

The previous subsections discussed the main aspects related to the reproducibility of MtL experiments. 
They refer to the alternatives and decisions taken that need to be properly reported.
Furthermore, some gaps were identified, mainly because it is unknown how the different choices could impact the characterization process. 
Below, each topic regarding the reproducible issues and gaps are summarized. 
The details can be seen in the respective subsection.

\begin{description}

	\item[Input domain:] Some measures support only categorical data while others, only numeric. 
	The alternatives to handle with this issue are \emph{ignoring}; \emph{transforming}, which implies in other decisions (see Figure~\ref{fig:measure-domain}); \emph{segmenting} the experiments and datasets.
	The impact of such choices in the statistical and information-theoretic meta-features is unknown. 
	Furthermore, datasets may have missing values, which will require imputation of values or the removal of the defective records.

	\item[Hyperparameters:] Some meta-features or groups of them require the definition of hyperparameters (see Table~\ref{tab:hyperparam}). 
	The way the hyperparameters affect the model-based and landmarking meta-features is unknown. 
	Also, approaches like tuning and the use of different hyperparameters values for the same measure have not been explored yet.

	\item[Range of the measures:] The meta-features have distinct range of values. 
	The alternatives to handle with this issue are \emph{ignoring} or \emph{transforming}. 
	In the latter (see Figure~\ref{fig:measure-range}), the \emph{min-max rescaling} and \emph{z-score normalization} are procedures that can be used; the \emph{dataset}, \emph{characterization measure} and the \emph{meta-feature} represent the objects to be transformed. 
	The gaps are concerned with identifying suitable combinations between the two dimensions and the normalization of the meta-features.

	\item[Summarization functions:] Different functions can be employed to summarize the measures result. 
	The investigation of how the summarization functions affect the measures' results are still incipient. 
	Furthermore, finding new alternatives to summarize the measures may increase the discriminative power of the meta-features.

	\item[Exceptions:] Some measures cannot be computed for all datasets. 
	The alternatives to handle this issue are \emph{ignoring} or \emph{replacing}. 
	In the latter, the alternatives are \emph{applying a preprocessing technique}; \emph{using a default value}; \emph{removing the missing values} (only for multi-valued measures). 
	However, the impact of such choices in the characterization result is unknown.
	
	\item[Meta-feature space:] Meta-feature dimensionality reduction can be performed using a \emph{feature-section} and/or \emph{transformation} approach. 
	In the former, \emph{wrapper} is more often used than \emph{filter}. 
	In the latter, although PCA is most commonly used, it is used in a small number of studies. 
	While the use of feature selection allows model interpretability, transformation usually has a lower computational cost.

\end{description}

We reinforce that many of those issues have not been properly reported in the MtL literature. 
This list can be used as a guideline for future studies involving dataset characterization. 
The next section addresses the characterization tools that contribute directly to reproducible empirical research in MtL. 

\section{Tools}
\label{sec:applications}

Characterization tools have an important role in the development of research in MtL. 
Besides simplifying an essential step of the work, their use corroborates the reproducibility of MtL experiments. 
However, the approach used in the development of the tool can generate two different perspectives: \emph{(i)} a black box tool with abstracted choices, which promotes reproducibility, but only for the users that use the same tool or, \emph{(ii)} a white box tool that exposes all the options to the user promoting reproducibility even with different tools, but forcing them to make the explicit decisions about the parameter values.

The Data Characterization Tool (DCT)\footnote{\url{https://github.com/openml/metafeatures/dct}} \citep{Lindner1999} is the most referenced characterization tool in the MtL literature~\citep{Bensusan2000a,Kopf2002,Pfahringer2000,Reif2014}, to cite a few. 
The DCT contains a representative subset of meta-features from simple, statistical and information-theoretic groups.

Matlab Statistics Toolbox~\citep{mathworks2001} have also been used to characterize statistical measures~\citep{Ali2006,Ali2006a,Smith-Miles2008}. Weka~\citep{Hall2009}, RapidMiner~\citep{Mierswa2006} and other general data mining tools can be employed to compute landmarking meta-features~\citep{Abdelmessih2010,Balte2014}. 

Nowadays, OpenML~\citep{OpenML2013} is the most robust tool available to characterize datasets, though it has a broader purpose. 
Many of the reported measures are available in the platform, which is also a benchmarking repository that contains the characterization of several datasets. 
OpenML uses an extension of the Fantail library~\citep{Sun2013}, also available on GitHub\footnote{\url{https://github.com/quansun/fantail-ml}, \url{https://github.com/openml/EvaluationEngine}}.
A drawback may be that the characterization process is performed automatically when a new dataset is submitted to the platform, which abstracts the users' choices. 
On the other hand, anyone can compute and upload their meta-features to OpenML through its API\footnote{\url{https://www.openml.org/api_docs\#!/data/post_data_qualities}}.

The framework proposed by~\citeA{Pinto2016} is available as an open GitHub project\footnote{\url{https://github.com/fhpinto/systematic-metafeatures}}, but without the implementation of the meta-features, which could be an expensive task.
Except for it, all the reviewed tools are black-box tools.


In parallel, many authors have used their implementation of the meta-features~\citep{Filchenkov2015,Garcia2015,Reif2014,Todorovski2000}, without reporting and making publicly available their implementation. 
This practice negatively affects reproducibly and comparison of results. 
Besides, without source code and widespread use, there is a chance that the implementations work as they should.
A positive step towards reproducibility is the \emph{``Paper with code"}\footnote{\url{https://paperswithcode.com/task/meta-learning/}} platform, which provides code repository.
However, comparing the  number of MtL related works published in the last 5 years with the number of codes available at the  \emph{``Paper with code"} website\footnote{Using Scopus, we found 412 papers related to MtL in the last 5 years, whereas we found only 65 works in the \emph{''Paper with code``} website.}, the practice of publishing the code/results is unfortunately still incipient.

\subsection{MFE Tool}
\label{subsec:mfe}

Aiming to offer a robust, flexible and standalone data characterization tool, the authors developed the Meta-Feature Extractor (MFE) tool\footnote{\new{Available in Python (\url{https://pypi.org/project/pymfe/}) and R (\url{https://cran.r-project.org/package=mfe}) languages}} that contains the implementation of most of the meta-features and summarization functions described in this paper. 
MFE also implements solutions for some of the issues discussed in Section \ref{sec:discussion} and provides a simple and flexible tool specifically designed to characterize datasets. 

MFE allows the user to compute a specific, a group of or all meta-features available.
It is possible to  define which summarization functions should be computed and, optionally, to obtain all computed values for a given set of measures, without summarizing the results. 
Many of the hyperparameters can be changed according to the user's preferences, as shown in Table~\ref{tab:hyperparam}, which also includes the default values adopted for all of them.
It is worth highlighting that the robustness of these choices, regarding the characterization process, is usually unknown, although they are consistent with the literature and the authors' experience.
The column ``Details" presents the rationale behind the decisions taken.

\begin{table}[ht]%
\footnotesize
\setlength{\tabcolsep}{4pt}
\begin{minipage}{\textwidth}
\begin{center}
\scalebox{1}{
\begin{tabularx}{\textwidth}{llcX}
  \toprule
  Measure & Hyperparameter & User & Details   \\
  \midrule
  
  \multicolumn{4}{l}{\emph{Statistical}} \\  
  all & transform = \texttt{TRUE} & Yes & Defined according to an exploratory analysis, to reduce the number of missing values in the meta-features.
  By setting it as \emph{true} the categorical attributes will be binarized using simple transformation, whereas with \emph{false} they will be ignored. \\ 
  & by.class = \texttt{FALSE} & Yes & Enables the measure extraction by class, as proposed by~\citeA{Castiello2005}. \\ \addlinespace
  
  \emph{cor} & method = ``pearson"  & Yes & Options: ``kendal" and ``spearman" \\ \addlinespace
  
  \emph{nrCorAttr} & method = ``pearson"  & Yes & Options: ``kendal" and ``spearman" \\
             & threshold = \texttt{0.5}  & No & As defined in~\citeA{Salama2013} \\ \addlinespace
  
  \emph{nrNorm} & W-Test for normality & No & Details in~\citeA{Royston1995} \\ 
  \emph{propNorm} & W-Test for normality & No & Details in~\citeA{Royston1995} \\ \addlinespace
  
  \emph{nrOutliers} & Tukey's boxplot & No & Details in~\citeA{Rousseeuw2011} \\
  \emph{propOutliers} & Tukey's boxplot & No & Details in~\citeA{Rousseeuw2011} \\ \addlinespace
  
  \emph{tMean} & trim = \texttt{0.2} & No & As defined in~\citeA{Ali2006a} \\ \addlinespace
  
\multicolumn{4}{l}{\emph{Information-theoretic}} \\
all & transform = \texttt{TRUE} & Yes & Defined according to an exploratory analysis, to reduce the number of missing values in the meta-features.
By setting it as \emph{true} the numeric attributes will be discretized using equal-frequency histogram transformation, whereas with \emph{false} they will be ignored. The number of bins is set to $\sqrt[3]{n}.$ \\  \addlinespace

\multicolumn{4}{l}{\emph{Model-based}} \\ 
all & algorithm = Cart & No & Details in~\citeA{Breiman1984}. \\ \addlinespace

\multicolumn{4}{l}{\emph{Landmarking}} \\ 
all & Cross-validation & No & Methodology used in order to obtain more stable results. \\
& folds = 10 & Yes & Also defines the measures cardinality. \\
& score = ``accuracy" & Yes & Options: ``balanced.accuracy" and ``kappa".\\ \addlinespace
bestNode & algorithm = Cart & No & Details in~\citeA{Breiman1984}. \\
randomNode & algorithm = Cart & No & Details in~\citeA{Breiman1984}. \\ 
worstNode & algorithm = Cart & No & Details in~\citeA{Breiman1984}. \\   \addlinespace

\multicolumn{4}{l}{\emph{Miscellaneous}} \\  
  
\emph{gravity} & distance = ``euclidian" & No & As defined in~\citeA{Ali2006}. \\
 
  \bottomrule
\end{tabularx}
}
\end{center}
\end{minipage}
\caption{Hyperparameters and their adopted default values in the MFE tool.}
\label{tab:hyperparam}
\end{table}%

As a limitation, MFE does not support to characterize non-classification datasets and does not accept datasets with missing values. 
An extension to other meta-features needs to follow the discussion described in Section \ref{sec:discussion}.
The authors believe that MFE can be used in any MtL experiment that requires the characterization of datasets, similar to DCT in the past, but with more flexibility.

\section{Conclusion}
\label{sec:conclusion}

The recommendation of techniques by using MtL is an effective alternative to deal with the selection of the most suitable techniques among a large number of possibilities.
However, many MtL studies adopt different methodologies and design approaches, which affect the reproducibility of the experiments. 
By discussing topics that have been frequently ignored in the MtL literature and suggesting possible alternatives to approach them, this paper reviewed the main characterization measures and important issues related to the reproducibility of MtL experiments,  in addition to the proposal of a new taxonomy for meta-features and the MFE tool.

The new taxonomy organized and formalized the current meta-features and their usefulness across different types of task, domain, range, and several other characteristics that can impact MtL tasks. 
Based on this review, the authors enumerated the main decisions a researcher faces when using meta-features.
Moreover, a detailed discussion is provided on the cutting edge subgroup of meta-features, their predictive power and the use cases where these measures have been applied.
In addition to this study, the MFE package was proposed to support the data characterization process implementing the framework proposed with the main meta-features included in the discussion.

Future work shall investigate meta-features for other types of tasks, such as regression and clustering; increase the interpretability of the meta-features; and explore empirical analysis showing how some choices related to the hyperparameters, cardinality and the summarization functions can affect dataset characterizations to best distinguish the performance of meta-models.
A review of regression and clustering meta-features could improve the task representation and could also look at a different perspective and validate the taxonomy proposed. 
The exploration of interpretability of the meta-features and the empirical analysis over hyperparameters, cardinality and summarization function could improve the meta-model representation and performance.

\section*{Acknowledgements}
This study was financed in part by the Coordena\c{c}\~{a}o de Aperfei\c{c}oamento de Pessoal de N\'{i}vel Superior - Brasil (CAPES) - Finance Code 001, CNPq (
152098/2016-0) and FAPESP (
2016/18615-0 and 2013/07375-0). The first, second and fifth authors would like to thank CeMEAI-FAPESP for the computational resources and Intel for the hardware and software server used in part of the experiments.

\appendix
\section{Characterization Measures Formalization}
\label{sec:formalization}

\subsection{Simple}

\begin{description}
\item[attrToInst] Ratio of the number of attributes per the number of instances~\citep{Kalousis1999}, also known as dimensionality: $\frac{d}{n}$.

\item[catToNum] Ratio of the number of categorical attributes per the number of numeric attributes~\citep{Feurer2014}: $\frac{\mathit{nrCat}_\textbf{X}}{\ {nrNum}_\textbf{X}}$.

\item[classToAttr] Ratio of the number of classes per the number of attributes~\citep{Todorovski2000}: $\frac{q}{d}$

\item[instToAttr] Ratio of the number of instances per the number of attributes~\citep{Kuba2002}: $\frac{n}{d}$.

\item[instToClass] Ratio of the number of instances per the number of classes~\citep{Vanschoren2010}: $\frac{n}{q}$.

\item[ntAttr] Number of attributes~\citep{Michie1994}: $d$.

\item[nrAttrMissing] Number of attributes with missing values~\citep{Feurer2014}: 
\begin{equation*}
\sum_{j=1}^{d}  \mathbbm{1}{(\sum_{i=1}^{n}\mathbbm{1}{(\vec{x}_{ij} = \varnothing)} > 0)}
\end{equation*}

\item[nrBin] Number of binary attributes \citep{Michie1994}: $\sum_{i=1}^{d} \mathbbm{1}{(\phi_{\vec{a}_{i}} = 2)}$. It includes numerical and categorical attributes that contain only two distinct values.

\item[nrCat] Number of categorical attributes \citep{Engels1998}: $d - \mathit{nrNum}_\textbf{X}$.

\item[nrClass] Number of classes \citep{Michie1994}: $q$.

\item[nrInst] Number of instances \citep{Michie1994}: $n$.

\item[nrInstMissing] Number of instances with missing values~\citep{Lindner1999}:
\begin{equation*}
\sum_{i=1}^{n}  \mathbbm{1}{(\sum_{j=1}^{d}\mathbbm{1}{(\vec{x}_{ij} = \varnothing)} > 0)}
\end{equation*}

\item[nrMissing] Number of missing values~\citep{Lindner1999}:
\begin{equation*}
\sum_{i=1}^{n} \sum_{j=1}^{d}\mathbbm{1}{(\vec{x}_{ij} = \varnothing)}
\end{equation*}

\item[nrNum] Number of numeric attributes \citep{Engels1998}: $\sum_{i=1}^{d}  \mathbbm{1}{(\vec{a}_{i} \in \mathds{R}^n)}$.

\item[numToCat] Ratio of the number of numeric attributes per the number of categorical attributes \citep{Feurer2014}: $\frac{\mathit{nrNum}_\textbf{X}}{\mathit{nrCat}_\textbf{X}}$.

\item[freqClass] Frequencies of the classes values \citep{Lindner1999}: \\$\left[\mathit{prop}_{c_1}, \ldots, \mathit{prop}_{c_q}\right]$, such that

\begin{equation}
\mathit{prop}_{c_j} = \frac{1}{n} \sum_{i=1}^{n} \mathbbm{1}{(y_i = c_j)}.
\label{eq:propclass}
\end{equation}

\end{description}

\subsection{Statistical}

\begin{description}
\item[canCor] Canonical correlations between the predictive attributes and the class \citep{Kalousis2002}: $\left[ \rho_1, \cdots, \rho_z \right]$, such that $\rho_i = \textit{cor}_{\vec{w}_x^{(i)}\mathbf{X},\vec{w}_y^{(i)}\mathbf{Y}}$, where $\vec{w}_x^{(i)}$ and $\vec{w}_y^{(i)}$ maximizes $\rho_i$ and are orthogonal to the $\vec{w}_x^{(i-1)}$ and $\vec{w}_y^{(i-1)}$, $\mathbf{Y}$ is the binarized version of $\vec{y}$ and $z \leq \mathit{min} \left[q, d\right]$ is the number of distinct $\vec{w}_x$ and $\vec{w}_y$ vectors found by using discriminant analysis. Frequently, the canonical correlation is reported in the literature as the eigenvalues of the canonical discriminant matrix, such that
\begin{equation}
\rho_i = \sqrt[]{\frac{\lambda_i}{1 + \lambda_i}}.
\label{eq:canCor}
\end{equation}

\item[cor] Absolute attributes correlation \citep{Castiello2005}:  $\left[ ~|cor_{\vec{a}_{1}, \vec{a}_{3}}|, \cdots, |cor_{\vec{a}_{d-1}, \vec{a}_{d}}|~ \right]$, such that $cor_{x,y}$ is obtained by the use of a correlation algorithm. The most common one used is the Pearson's Correlation coefficient, given by
\begin{equation}
cor_{x,y} = \frac{\mathit{cov}_{x,y}}{\mathit{sd}_x\mathit{sd}_y}\textrm{, where}
\label{eq:cor}    
\end{equation} 
\begin{equation}
cov_{x,y}=\frac{\sum_{i=1}^{n}(x_{i}-\bar{x})(y_{i}-\bar{y})}{n-1}\textrm{, and}
\label{eq:cov}
\end{equation}
\begin{equation}
sd_{x}=\sqrt[2]{\frac{\sum^n_{i=1}(x_i - \bar{x})^2}{n - 1}}
\label{eq:sd}
\end{equation}

\item[cov] Attributes covariance \citep{Castiello2005}: $\left[ ~|cov_{\vec{a}_{1}, \vec{a}_{2}}|, \cdots, |cov_{\vec{a}_{d-1}, \vec{a}_{d}}|~\right]$, where $cov_{x,y}$ is given by Equation \ref{eq:cov}.

\item[nrDisc] Number of discriminant functions \citep{Lindner1999}: $|\mathit{canCor}_{\mathcal{D}}|$.

\item[eigenvalues] Eigenvalues of the covariance matrix \citep{Ali2006}: $\left[ \lambda_1, \cdots, \lambda_d \right]$, such that $S\vec{v} = \lambda_i\vec{v}$ for some $\vec{v} \neq 0$, where $S_{d{\times}d}$ is the covariance matrix of $\textbf{X}$.

\item[gMean] Geometric mean of attributes \citep{Ali2006a}: $\left[\mathit{gMean}_{\vec{a}_{1}}, \cdots,  \mathit{gMean}_{\vec{a}_{d}} \right]$, such that
$\mathit{gMean}_x = \Big(\prod^n_{i=1} x_i\Big)^{\frac{1}{n}}$.

\item[hMean] Harmonic mean of attributes \citep{Ali2006a}: $\left[\mathit{hMean}_{\vec{a}_{1}}, \cdots,  \mathit{hMean}_{\vec{a}_{d}} \right]$, such that \[ \mathit{hMean}_x = \frac{n}{\sum^{n}_{i=1} \frac{1}{x_i}}. \]

\item[iqRange] Interquartile range of attributes \citep{Ali2006a}: $\left[\mathit{iqRange}_{\vec{a}_{1}}, \cdots,  \mathit{iqRange}_{\vec{a}_{d}} \right]$, such that $\mathit{iqRange}_x = \mathit{Q3}_x - \mathit{Q1}_x$, where $\mathit{Q1}_x$ and $\mathit{Q3}_x$ represent the first and third quartile values of $x$, respectively.

\item[kurtosis] Kurtosis of attributes \citep{Michie1994}: $\left[\mathit{kurt}_{\vec{a}_{1}}, \cdots,  \mathit{kurt}_{\vec{a}_{d}} \right]$, such that \[\mathit{kurt}_x = \frac{m_4}{\textit{sd}_x^4} - 3,\] where $m_j$ represents a statistical moment, given by
\begin{equation}
m_j = \frac{1}{n} \sum^n_{i=1} (x_i - \bar{x})^j.
\label{eq:moment}
\end{equation}

\item[mad] Median absolute deviation of attributes \citep{Ali2006}: $\left[\mathit{mad}_{\vec{a}_{1}}, \cdots, \mathit{mad}_{\vec{a}_{d}} \right]$, such that $\mathit{mad}_x = \mathit{median} \left[ ~|x_1 - \mathit{median}_{x}|, \cdots, |x_n - \mathit{median}_{x}|~ \right]$, where 
\begin{equation}
\mathit{median}_x =
\begin{cases}
      \frac{1}{2}(x_{(r)} + x_{(r+1)})
    & \text{if $|x|$ is even $(|x| = 2r)$}
    \\[4pt]
      x_{(r+1)}
      & \text{otherwise $(|x| = 2r+1)$}
\end{cases}
\label{eq:median}
\end{equation}

\item[max] Maximum value of attributes~\citep{Engels1998}: $\left[\mathit{max}~{\vec{a}_{1}}, \cdots, \mathit{max}~{\vec{a}_{d}} \right]$.

\item[mean] Mean value of attributes~\citep{Engels1998}: $\left[~\overline{\vec{a}_{1}}, \cdots, \overline{\vec{a}_{d}} ~\right]$.

\item[median] Median value of attributes~\citep{Engels1998}: \\$\left[\mathit{median}_{\vec{a}_{1}}, \cdots,  \mathit{median}_{\vec{a}_{d}} \right]$, where $\mathit{median}_{x}$ is given by Equation \ref{eq:median}.

\item[min] Minimum value of attributes~\citep{Engels1998}: $\left[\mathit{min}~{\vec{a}_{1}}, \cdots, \mathit{min}~{\vec{a}_{d}} \right]$.

\item[nrCorAttr] Number of attributes pairs with high correlation~\citep{Salama2013}:
\[ \frac{2}{d(d-1)} \sum^{d-1}_{i=1}\sum^d_{j=i+1} \mathbbm{1}{(|\mathit{cor}_{\vec{a}_{i}, \vec{a}_{j}}| \geq \tau)},\]
where $\tau$ is a threshold value between $0$ and $1$, usually $\tau = 0.5$.  This is the normalized version adapted by the authors.

\item[nrNorm] Number of attributes with normal distribution~\citep{Kopf2000}: \\$\sum^d_{i=1} \mathbbm{1}{(\mathit{isNormal}_{\vec{a}_{i}})}$. To check if an attribute has or does not have a normal distribution the W-Test for normality~\citep{Royston1995} can be applied, for instance.

\item[nrOutliers] Number of attributes with outliers values~\citep{Kopf2002}: \\$\sum^d_{i=1} \mathbbm{1}(\mathit{hasOutlier}_{\vec{a}_{j}})$. To test if an attribute has or does not have outliers, the Tukey's boxplot algorithm~\citep{Rousseeuw2011} can be used, for instance.

\item[range] Range of Attributes~\citep{Ali2006a}: \\$\left[(\mathit{max}~\vec{a}_{1} - \mathit{min}~\vec{a}_{1}), \cdots,  (\mathit{max}~\vec{a}_{d} - \mathit{min}~\vec{a}_{d}) \right]$.

\item[sd] Standard deviation of the attributes~\citep{Engels1998}: $\left[\mathit{sd}_{\vec{a}_{1}}, \cdots,  \mathit{var}_{\vec{a}_d}\right]$, such that $\mathit{sd}_x$ is given by Equation \ref{eq:sd}.

\item[sdRatio] Statistic test for homogeneity of covariances~\citep{Michie1994}:
\[\begin{split}
\mathit{exp}(M/d \sum^q_{i=1}(n_{c_i} - 1))\text{, where } 
M = \gamma\sum^q_{i=1}(n_{c_i}-1)\mathit{log} |S_i^{-1}S|\text{;}\\
\gamma = 1 - \frac{2d^2+3d-1}{6(d+1)(q-1)} \sum^q_{i=1}\frac{1}{n_{c_i} - 1}-\frac{1}{n-q}\text{;}\\
S = \frac{1}{n - q}\sum^q_{i=1}(n_{c_i}-1)S_i
\end{split}\]
such that, $n_{c_i}$ is the number of instances related to the class $c_i$, $S$ is called pooled covariance matrix and $S_i$ is the sample covariance matrix of the instances for the $i^{th}$ class.

\item[skewness] Skewness of attributes~\citep{Michie1994}: $\left[\mathit{skewness}_{\vec{a}_{1}}, \cdots,  \mathit{skewness}_{\vec{a}_{d}} \right]$, such that \[\mathit{skewness}_x = \frac{ m_3 }{ \mathit{sd}_x^3 },\] where $\mathit{sd}_x$ and $m_3$ are given by Equation \ref{eq:sd} and \ref{eq:moment}, respectively.

\item[tMean] Trimmed mean of attributes~\citep{Engels1998}: $\left[\mathit{tMean}_{\vec{a}_{1}}, \cdots,  \mathit{tMean}_{\vec{a}_{d}} \right]$, such that
\[\mathit{tMean}_x = \frac{x_{(i+1)} + x_{(i+2)} + \cdots + x_{(n-i-2)} + x_{(n-i-1)}}{n - 2i}, \]
where $i = \lceil n \alpha \rceil$ and $\alpha$ is a hyperparameter, such that $0 < \alpha < 0.5$. The suggested value is $\alpha = 0.2$.

\item[var] Attributes variance~\citep{Castiello2005}: $\left[\mathit{var}_{\vec{a}_{1}}, \cdots,  \mathit{var}_{\vec{a}_d}\right]$, such that 
\[\mathit{var}_{x} = \frac{\sum^n_{i=1}(x_i - \bar{x})^2}{n - 1}.\]

\item[wLambda] Wilks Lambda~\citep{Lindner1999}: \[\prod_{i=1}^z \frac{1}{1+\lambda_i},\] where $z = \mathit{nrDisc}_{\mathcal{D}}$ and $\lambda_i$ is defined in Equation \ref{eq:canCor}.

\end{description}
 
\subsection{Information-Theoretic}
Let $H_x$ be the entropy of a given attribute, such that
\begin{equation*}
H_x = -\sum^{\phi_x}_{i=1} P(x = \varphi^x_i) \log_2 P(x = \varphi^x_i)\text{,}
\end{equation*}
and let $\mathit{H}_{x,y}$ be the joint entropy of a predictive attribute $x$ and the class $y$, such that
\begin{equation*}
\mathit{H}_{x,y} = \sum^{\phi_x}_{i=1}\sum^{\phi_y}_{j=1} \pi_{ij} \log_2 \pi_{ij}\text{,}
\end{equation*} where $\pi_{ij} = P(x = \varphi^x_i, y = \varphi^y_j)$. The mutual information shared between them is given by
$\mathit{MI}_{x,y} = H_x + H_y - H_{x,y}$. Mainly from these concepts, the information-theoretic measures are computed as following:

\begin{description}

\item[attrEnt] Attributes entropy~\citep{Michie1994}: $\left[H_{\vec{a}_{1}}, \cdots,  H_{\vec{a}_{d}} \right]$.

\item[classEnt] Class entropy~\citep{Michie1994}: $H_{\vec{y}}$

\item[eqNumAttr] Equivalent number of attributes~\citep{Michie1994}: 
\[\frac{H_{\vec{y}}}{\frac{1}{d} \sum^d_{i=1} \textit{MI}_{\vec{a}_{i}, \vec{y}}}\]

\item[jointEnt] Joint Entropy of attributes and classes~\citep{Michie1994}: \\$\left[H_{\vec{a}_{1}, \vec{y}}, \cdots,  H_{\vec{a}_{d}, \vec{y}} \right]$.

\item[mutInf] Mutual information of attributes and classes~\citep{Michie1994}: \\$\left[\textit{MI}_{\vec{a}_{1}, \vec{y}}, \cdots,  \textit{MI}_{\vec{a}_{d}, \vec{y}} \right]$.

\item[nsRatio] Noisiness of attributes~\citep{Michie1994}	:
\[\frac{\frac{1}{d}\sum^d_{j=1} H_{\vec{a}_j} - \frac{1}{d}\sum^d_{j=1} \mathit{MI}_{\vec{a}_j, \vec{y}}}{\frac{1}{d}\sum^d_{j=1} \mathit{MI}_{\vec{a}_j, \vec{y}}}.\]

\end{description}

\subsection{Model-Based}
For DT-model meta-features, let $\psi$ be the set of leaves, $\eta$ be the set of nodes, such that $\psi \cap \eta = \emptyset$ and $\Gamma = \psi \cup \eta$ are the whole structure of the tree that represents the DT learning model. In addition, consider the following tree properties:
\begin{description}
\item[$\mathit{attr}_{\eta_i}$] Predictive attribute used in the node $\eta_i$.
\item[$\mathit{class}_{\psi_i}$] Class predicted by the leaf $\psi_i$.
\item[$\mathit{inst}_{\Gamma_i}$] Number of training instances used to define the tree element $\Gamma_i$.
\item[$\mathit{level}_{\Gamma_i}$] Level of the tree element $\Gamma_i$. In other words, it is the number of nodes in the tree hierarchy necessary to reach the root of the tree, such that $\mathit{level}_{\Gamma_i} = 0 ~\mathrm{iff}~ {\Gamma_i} = \mathit{root}_\Gamma$.
\item[$\mathit{prob}_{\psi_i}$] Probability of reaching the leaf $\psi_i$ from the root in a random walk through the tree hierarchy, such that $\mathit{prob}_{\psi_i} = \frac{1}{2^{\mathit{level}_{\psi_i}}}$.
\item[$\mathit{root}_\Gamma$] Root node of a tree, such that $\mathit{root}_\Gamma \in \eta$.
\end{description}

The DT-model meta-features are the following:

\begin{description}
\item[leaves] Number of leaves~\citep{Peng2002a}: $|\psi|$.

\item[leavesBranch] Size of branches~\citep{Peng2002a}: $\left[\mathit{level}_{\psi_1}, \cdots, \mathit{level}_{\psi_z}) \right]$, where $z = |\psi|$.

\item[leavesCorrob] Leaves corroboration~\citep{Bensusan2000}: $\left[\frac{\mathit{inst}_{\psi_1}}{n}, \cdots, \frac{\mathit{inst}_{\psi_{z}}}{n} \right]$, where $z = |\psi|$.

\item[leavesHomo] Homogeneity~\citep{Bensusan2000}: $\left[\frac{z}{\mathit{shape}_{\psi_1}}, \cdots, \frac{z}{\mathit{shape}_{\psi_z}} \right]$, where $z = |\psi|$.

\item[leavesPerClass] Leaves per class~\citep{Filchenkov2015}: $\left[ \mathit{lpc}_{c_1}, \cdots, \mathit{lpc}_{c_q} \right]$, such that
\[\mathit{lpc}_{c_j} = \frac{1}{|\psi|} \sum^{|\psi|}_{i=1} \mathbbm{1}{(\mathit{class}_{\psi_i} = c_j)} \]

\item[nodes] Number of nodes~\citep{Peng2002a}: $|\eta|$.

\item[nodesPerAttr] Ratio of the number of nodes per the number of attributes~\citep{Bensusan2000}: $\frac{|\eta|}{d}$.

\item[nodesPerInst] Ratio of the number of nodes per the number of instances~\citep{Bensusan2000}: $\frac{|\eta|}{n}$.

\item[nodesPerLevel] Number of nodes per level~\citep{Peng2002a}: $\left[ \mathit{npl}_1, \cdots, \mathit{npl}_{\mathit{level}_w} \right]$, such that 
\[w = \operatorname*{arg\,max}_{\eta_i \in \eta} \mathit{level}_{\eta_i}\mathrm{, and}\]
\[\mathit{npl}_j = \sum^{|\eta|}_{i=1} \mathbbm{1}{(\mathit{level}_{\eta_i} = j)}.\]

\item[nodesRepeated] Repeated nodes~\citep{Bensusan2000}: $\left[\mathit{nrp}_1, \cdots, \mathit{nrp}_d \right] ~\forall~ \mathit{nrp}_j > 0$, such that
\[\mathit{nrp}_j = \sum^{|\eta|}_{i=1} \mathbbm{1}{(\mathit{attr}_{\eta_i} = j)}.\]

\item[treeDepth] Tree depth~\citep{Peng2002a}: $\left[\mathit{level}_{\Gamma_1}, \cdots, \mathit{level}_{\Gamma_{w}} \right]$, where $w = |\Gamma|$.

\item[treeImbalance] Tree imbalance~\citep{Bensusan2000}: $\left[ \mathit{imb}_{\psi_1}, \cdots, \mathit{imb}_{\psi_w} \right]$, such that $w = {|\psi|}$ and $\mathit{imb}_{\psi_j} =  -z_{\psi_j} (log_2 ~z_{\psi_j})$, where $z_{\psi_j} = \mathit{prob}_{\psi_j} \sum_{i=1}^w \mathbbm{1}{(\mathit{prob}_{\psi_i} = \mathit{prob}_{\psi_j})}$.

\item[treeShape] Tree shape~\citep{Bensusan2000}: $\left[ \mathit{shape}_{\psi_1}, \cdots, \mathit{shape}_{\psi_w} \right]$, such that $\mathit{shape}_{\psi_j} = -\mathit{prob}_{\psi_j} (log_2 ~ \mathit{prob}_{\psi_j})$ and $w = {|\psi|}$.

\item[varImportance] Variable importance~\citep{Bensusan2000}: $\left[ \mathit{imp}_{\vec{a}_{1}, \vec{y}},  \cdots, \mathit{imp}_{\vec{a}_{d}, \vec{y}} \right]$, where $\mathit{imp}_{\vec{a}_{j}, \vec{y}}$ describes the homogeneity of the class $\vec{y}$ produced by some split of a given attribute $\vec{a}_{j}$. Each DT learning algorithm uses a specific procedure to define the importance of the variables.
\end{description}

\subsection{Landmarking}
Let $\mathcal{A}$, $\theta_A$ and $\xi$ be, respectively, a learning algorithm, a learning model and an evaluation measure. All landmarking meta-features are computed in the same way, such as the model is induced using the learning algorithm and a train data:
\[ \mathcal{A}(\mathbf{X}_{train}, \vec{y}_{train}) \to \theta_A \]
and the prediction of the learning model for a test data is evaluated using the given measure, such as
\[ \mathit{landmarking}_A = \xi(\theta_A(\mathbf{X}_{test}), \vec{y}_{test})\text{,} \]
where \emph{train} and \emph{test} subset are defined for each fold.

The differences between the landmarking measures are given by the learning-algorithm family and the predictive attributes used to induce the model, as described below:

\begin{description}
\item[bestNode] Decision Node: $\mathcal{A}_\mathrm{DT}(\mathbf{X}_{train,battr}, ~\vec{y}_{train}) \to \theta_\mathrm{bestNode}$, where $\mathbf{X}_{train,battr}$ is the content of the most informative attribute, which is defined using the \emph{varImportance} result.

\item[eliteNN] Elite Nearest Neighbor: $\mathcal{A}_\mathrm{KNN}(\mathbf{X}_{train,battrset}, ~\vec{y}_{train}, ~k=1) \to \theta_\mathrm{eliteNN}$, where $\mathbf{X}_{train,battrset}$ contains only the subset of the most informative attributes for the train data. They are defined using the \emph{varImportance} result.

\item[linearDiscr] linear Discriminant: $\mathcal{A}_\mathrm{LD}(\mathbf{X}_{train}, ~\vec{y}_{train}) \to \theta_\mathrm{linearDiscr}$.

\item[naiveBayes] Naive Bayes: $\mathcal{A}_\mathrm{NB}(\mathbf{X}_{train}, ~\vec{y}_{train}) \to \theta_\mathrm{naiveBayes}$.

\item[oneNN] One Nearest Neighbor: $\mathcal{A}_\mathrm{KNN}(\mathbf{X}_{train}, ~\vec{y}_{train}, ~k=1) \to \theta_\mathrm{oneNN}$.

\item[randomNode] Random node: $\mathcal{A}_\mathrm{DT}(\mathbf{X}_{train,rattr}, ~\vec{y}_{train}) \to \theta_\mathrm{randomNode}$, where $\mathbf{X}_{train,rattr}$ is the content of a random attribute.

\item[worstNode] Worst node: $\mathcal{A}_\mathrm{DT}(\mathbf{X}_{train,wattr}, ~\vec{y}_{train}) \to \theta_\mathrm{worstNode}$, where $\mathbf{X}_{train,wattr}$ is the content of the least informative attribute.

\end{description}

\subsection{Others}
The following subsections specify the non-traditional characterization measures, that include groups and standalone meta-features.

\subsubsection{Clustering and distance-based}
The clustering and distance-based measures use the result of a clustering algorithm and/or a distance measure.
The $k$ partitions obtained from the use of a clustering algorithm are denoted by $C_i \subset \mathcal{D}$, such that $\overline{\vec{x}}_{C_i}$ denotes the center of cluster $i$. 
Without loss of generality, $\mathit{dist}_{x,y}$ represents a distance between two instances  $\vec{x}_i \in \mathcal{D},~ \vec{x}_j \in \mathcal{D}$, regardless of the type of attributes they have.

\begin{description}
\item[AIC] Akaike Information Criterion~\citep{Vukicevic2016}:
\begin{equation*}
    \sum^k_{i=1}\sum_{\vec{x}_j \in C_i} (\vec{x}_j - \overline{\vec{x}}_{C_i})^2 + 2dk.
\end{equation*}

\item[BIC] Bayesian Information CriterionVukicevic2016
\begin{equation*}
    \sum^k_{i=1}\sum_{\vec{x}_j \in C_i} (\vec{x}_j - \overline{\vec{x}}_{C_i})^2 + dk~\mathit{log}_n.
\end{equation*}

\item[compactness] Quantify the compactness of the partitions \citep{Vukicevic2016}: $\left[c_1, \cdots, c_k \right]$, such that
\begin{equation}
    c_i = \sum_{\vec{x}_j \in C_i} \mathit{dist}_{\vec{x}_j,\overline{\vec{x}}_{C_i}}.
\label{eq:compactness}
\end{equation}

\item[connectivity] Amount of neighbouring instances that are not in the same partition~\citep{Vukicevic2016}:
\begin{equation*}
    \sum^n_{i=1} \mathbbm{1}{(\vec{x}_j \in C_i \wedge nn_{\vec{x}_i} \not\in C_i)},
\end{equation*}
where $nn_{\vec{x}_i}$ is the nearest neighbor of instance $\vec{x}_i$.

\item[distInst] Distance between all pairs of instances~\cite{Ferrari2015}: \begin{equation}
\left[ \textit{dist}_{\vec{x}_1,\vec{x}_2}, \textit{dist}_{\vec{x}_1,\vec{x}_3}, \cdots, \textit{dist}_{\vec{x}_{n-2},\vec{x}_n}, \textit{dist}_{\vec{x}_{n-1},\vec{x}_n} \right]
\label{eq:distInst}
\end{equation}

\item[distCorrInst] Distance and correlations of all pairs of instances~\cite{Pimentel2019}: $\left[c', d'\right]$, where
\begin{equation*}
\begin{gathered}
c' =  \frac{c + 1}{2},\qquad c = \left[ \textit{cor}_{\vec{x}_1,\vec{x}_2}, \textit{cor}_{\vec{x}_1,\vec{x}_3}, \cdots, \textit{cor}_{\vec{x}_{n-2},\vec{x}_n}, \textit{cor}_{\vec{x}_{n-1},\vec{x}_n} \right], \\
d' = \frac{d - \min(d)}{\max(d) - \min(d)},
\end{gathered}
\end{equation*}
such that $\mathit{cor}_{x,y}$ (Equation~\ref{eq:cor}) is used to compute the correlation between 2 instances and $d$ is given by Equation~\ref{eq:distInst}. 

\item[gravity] Center of gravity~\citep{Ali2006}: $\mathit{dist}_{\overline{\vec{x}}_{C_{m}}, \overline{\vec{x}}_{C_{n}}}$, where $\overline{\vec{x}}_{C_{m}}$ and $\overline{\vec{x}}_{C_{n}}$ are the center points of the instances related to the majority and minority classes, respectively.

\item[nrClusters] Number of clusters~\citep{Nascimento2009}: $|C| = k$.

\item[purityRatio] Ratio of the number of clusters with a given class \citep{Ler2018}: $\left[\frac{s_1}{k}, \cdots, \frac{s_q}{k} \right]$, where
\begin{equation*}
s_i = \sum^k_{j=1} \mathbbm{1}{((\vec{x}_l, y_i) \in C_j)}
\end{equation*}
\item[silhouette] Global silhouette index~\citep{Vukicevic2016}:
\begin{equation*}
\begin{gathered}
    \frac{1}{k} \sum^k_{i=1} \Big( \frac{1}{|C_i|} \sum_{\vec{x}_j \in C_i} \frac{b(\vec{x}_j) - a(\vec{x}_j)}{max(a(\vec{x}_j), b(\vec{x}_j))} \Big), \mathrm{where}\\
    a(\vec{x}_j) = \frac{1}{|C_i| - 1} \sum_{\substack{\vec{x}_l \in C_i\\j \not= l}} \mathit{dist}_{\vec{x}_j, \vec{x}_l},\qquad
    b(\vec{x}_j) =  {\min_{i \not= i'}}\Big(\frac{1}{|C_{i'}|} \sum_{\vec{x}_l \in C_{i'}} \mathit{dist}_{\vec{x}_j, \vec{x}_l}\Big).
\end{gathered}
\end{equation*}

\item[sizeDist] Proportion of instances present in each cluster~\citep{Ler2018}: $\left[\frac{|C_1|}{n}, \cdots, \frac{|C_k|}{n} \right]$.

\item[XB] Xie-Beni index~\citep{Vukicevic2016}:
\begin{equation*}
\begin{gathered}
\frac{1}{n}\frac{\sum^k_{i=0} c_i }{\smash{\displaystyle\min_{i<i'}}~\delta(C_i, C_{i'})}, \qquad
\delta(C_i, C_{i'}) = \min_{\substack{\vec{x}_j \in C_i\\\vec{x}_l \in C_{i'}}} \mathit{dist}_{\vec{x}_j, \vec{x}_l},
\end{gathered}
\end{equation*}
where $c_i$ is given by Equation~\ref{eq:compactness}.

\end{description}

\subsubsection{Complexity Measures}
The complexity measures are specified, well described and explained in \citet{Lorena2018}.

\subsubsection{Miscellaneous}
\begin{description}
\item[attrConc] Attributes concentration coefficient~\citep{Kalousis2001a}: \\$\left[\mathit{conc}_{\vec{a}_{1}, \vec{a}_{2}}, \mathit{conc}_{\vec{a}_{2}, \vec{a}_{1}}, \cdots, \mathit{conc}_{\vec{a}_{d-1}, \vec{a}_{d}}, \mathit{conc}_{\vec{a}_{d}, \vec{a}_{d-1}} \right]$, such that
\begin{equation}
\begin{split}
\mathit{conc}_{x,y} = \frac{\sum^{\phi_x}_{i=1}\sum^{\phi_y}_{j=1} \frac{\pi_{ij}^2}{\pi_{i+}} - \sum^{\phi_y}_{j=1} \pi^2_{+j}}{1 - \sum^{\phi_j}_{j=1} \pi^2_{+j}}, \mathrm{where} \\
\pi_{ij} = P(x = \varphi^x_i, y = \varphi^y_j),\quad 
\pi_{i+} = \sum^{\phi_y}_{j=1} \pi_{ij} \quad
\mathrm{and}\quad \pi_{+j} = \sum^{\phi_x}_{i=1} \pi_{ij}.
\end{split}
\label{eq:concentration}
\end{equation}

\item[classConc] Class concentration coefficient~\citep{Kalousis2001a}: \\$\left[\mathit{conc}_{\vec{a}_{1}, \vec{y}}, \cdots, \mathit{conc}_{\vec{a}_{d}, \vec{y}} \right]$, where $\mathit{conc}_{x,y}$ is given by Equation \ref{eq:concentration}.

\item[cohesiveness] Density of the example distribution~\citep{Vilalta2002a}: $\left[v({\vec{x}_1}), \cdots, v({\vec{x}_n}) \right]$.
\begin{equation*}
    v({\vec{x}_i}) = \frac{1}{|\mathcal{K}|} \sum_{(\vec{x}_j, y_j) \in \mathcal{K}_{\vec{x}_i}} 1 - \mathbbm{1}{(y_i = y_j)},  
    \label{eq:cohesiveness}
\end{equation*}
where $\mathcal{K}_{\vec{x}_i}$ contains the $k$ nearest neighbors of instance $\vec{x}_i$. The $k$ is a user hyperparameter.

\item[consistencyRatio] Proportion of repeated instances that have different targets~\citep{Kopf2002}:
\begin{equation*}
    \frac{1}{n} \sum^n_{i=2} \sum^{i-1}_{j=1} \mathbbm{1}{(\mathit{dist}_{\vec{x}_i,\vec{x}_j} = 0 \wedge y_i \not= y_j)} 
\end{equation*}

\item[incoherenceRatio] Ratio of instances that does not overlap with any other instances in a predefined number of attributes~\citep{Kopf2002}:
\begin{equation*}
    \frac{1}{n} \sum^n_{i=2} \mathbbm{1}{\Big( \sum^{i-1}_{j=1} \mathbbm{1}{(o(\vec{x}_i, \vec{x}_j) > \alpha}) = 0 \Big)} \qquad
    o(\vec{x}_i, \vec{x}_j) = \sum^d_{l=1} \mathbbm{1}{(v_{il} = v_{jl})},
\end{equation*}
where $\alpha$ is a user hyperparameter to set the number of similar attributes to define when two instances overlap.

\item[infotheoTime] The elapsed time to compute the information theretical meta-features~\citep{Reif2011}.

\item[landTime] The elapsed time to compute the landmarkings meta-features~\citep{Reif2011}.

\item[modelTime] The elapsed time to compute the model-based meta-features~\citep{Reif2011}.

\item[oneItemset] Frequency of the predictive attributes after they are binarized~\citep{Song2012}: 
\begin{equation*}
    \left[\frac{\sum^n_{i=1} v_{i1}}{n}, \cdots, \frac{\sum^n_{i=1} v_{id}}{n}) \right].
\end{equation*}

\item[propPCA] Proportion of principal components that explain a specific variance of the dataset~\citep{Feurer2014}: 
\begin{equation*}
    \frac{|\Lambda| - \sum^{|\lambda|}_{i=1} \mathbbm{1}{\Big(\sum^{i}_{j=1} \lambda_i > \alpha\Big)} + 1}{|\Lambda|},
\end{equation*}
where $\Lambda$ is the set of all eigen values $\lambda_i$ inversely ordered according to their variance and $\alpha$ is a user defined threshold indicating the amount of variance desired, e.g. 0.95. 
 

\item[sparsity] Attributes sparsity~\citep{Salama2013}: $\left[\mathit{sparsity}_{\vec{a}_{1}}, \cdots,  \mathit{sparsity}_{\vec{a}_{d}} \right]$, such that
\[ \mathit{sparsity}_x = \frac{1}{n-1} \Big{(} \frac{\sum_{i=1}^{\phi_{x}} N(x = \varphi^{x}_i)}{\phi_{x}} - 1\Big{)},\]
where $N(x = \varphi^{x}_i)$ is the number of times that the $i^{th}$ distinct value of $x$ are present in the vector. This is the normalized version adapted by the authors.

\item[statTime] The elapsed time to compute the statistical meta-features~\citep{Reif2011}.

\item[twoItemset] Frequency of predictive attributes' pairs after they are binarized~\citep{Song2012}:\\ $\left[v(1, 2), v(1, 3), \cdots, v(d-2, d), v(d-1, d)) \right]$
\begin{equation*}
    v(i, j) = \frac{1}{n} \sum^n_{l=1} \mathbbm{1}{(v_{li} \not= v_{lj})}.
\end{equation*}

\item[uniquenessRatio] Proportion of repeated instances~\citep{Kopf2002}:
\begin{equation*}
    \frac{1}{n} \sum^n_{i=2} \sum^{i-1}_{j=1} \mathbbm{1}{(\mathit{dist}_{\vec{x}_i,\vec{x}_j} = 0)} 
\end{equation*}

\item[wgDist] Weighted distance~\citep{Vilalta1999}: $\left[v(\vec{x}_1), \cdots, v(\vec{x}_n) \right]$, where
\begin{equation*}
\begin{gathered}
    v(\vec{x}_i) = \frac{\sum^n_{j=1,j\not=i} W(\vec{x}_i, \vec{x}_j) \mathit{dist}_{\vec{x}_i, \vec{x}_j}}{\sum^n_{j=1,j\not=i} W(\vec{x}_i, \vec{x}_j)} \\
    W(\vec{x}_i, \vec{x}_j) = \frac{1}{2^{2d}} \qquad d = \frac{\mathit{dist}_{\vec{x}_i, \vec{x}_j}}{\sqrt{n-\mathit{dist}_{\vec{x}_i, \vec{x}_j}}}
\end{gathered}
\end{equation*}

\end{description}

\bibliographystyle{elsarticle-harv} 
\bibliography{main}

\begin{thebibliography}{88}
\expandafter\ifx\csname natexlab\endcsname\relax\def\natexlab#1{#1}\fi
\expandafter\ifx\csname url\endcsname\relax
  \def\url#1{\texttt{#1}}\fi
\expandafter\ifx\csname urlprefix\endcsname\relax\def\urlprefix{URL }\fi

\bibitem[{Abdelmessih et~al.(2010)Abdelmessih, Shafait, Reif, and
  Goldstein}]{Abdelmessih2010}
Abdelmessih, S.~D., Shafait, F., Reif, M., Goldstein, M., 2010. Landmarking for
  meta-learning using {RapidMiner}. In: RapidMiner Community Meeting and
  Conference (RCOMM). pp. 1--6.

\bibitem[{Aggarwal(2015)}]{Aggarwal2015}
Aggarwal, C.~C., 2015. Data Mining. Springer International Publishing.

\bibitem[{Ali and Smith(2006)}]{Ali2006}
Ali, S., Smith, K.~A., 2006. On learning algorithm selection for
  classification. Applied Soft Computing 6~(2), 119 -- 138.

\bibitem[{Ali and Smith-Miles(2006)}]{Ali2006a}
Ali, S., Smith-Miles, K.~A., 2006. A meta-learning approach to automatic kernel
  selection for support vector machines. Neurocomputing 70~(1), 173 -- 186.

\bibitem[{Balte et~al.(2014)Balte, Pise, and Kulkarni}]{Balte2014}
Balte, A., Pise, N., Kulkarni, P., 2014. Meta-learning with landmarking : {A}
  {S}urvey. International Journal of Computer Applications 105~(8), 47 -- 51.

\bibitem[{Barella et~al.(2018)Barella, Garcia, de~Souto, Lorena, and
  de~Carvalho}]{Barella2018}
Barella, V.~H., Garcia, L. P.~F., de~Souto, M. C.~P., Lorena, A.~C.,
  de~Carvalho, A. C. P. L.~F., 2018. Data complexity measures for imbalanced
  classification tasks. In: 2018 International Joint Conference on Neural
  Networks, {IJCNN} 2018, Rio de Janeiro, Brazil, July 8-13, 2018. {IEEE}, pp.
  1--8.

\bibitem[{Bensusan and Giraud-Carrier(2000)}]{Bensusan2000a}
Bensusan, H., Giraud-Carrier, C., 2000. Discovering task neighbourhoods through
  landmark learning performances. In: 4th European Conference on Principles of
  Data Mining and Knowledge Discovery (PKDD). pp. 325 -- 330.

\bibitem[{Bensusan et~al.(2000)Bensusan, Giraud-Carrier, and
  Kennedy}]{Bensusan2000}
Bensusan, H., Giraud-Carrier, C., Kennedy, C., 2000. A higher-order approach to
  meta-learning. In: 10th International Conference Inductive Logic Programming
  (ILP). pp. 33 -- 42.

\bibitem[{Bensusan and Kalousis(2001)}]{Bensusan2001}
Bensusan, H., Kalousis, A., 2001. Estimating the predictive accuracy of a
  classifier. In: 12th European Conference on Machine Learning (ECML). pp. 25
  -- 36.

\bibitem[{Bilalli et~al.(2017)Bilalli, Abell{\'{o}}, and
  Aluja-Banet}]{Bilalli2017}
Bilalli, B., Abell{\'{o}}, A., Aluja-Banet, T., 2017. On the predictive power
  of meta-features in {OpenML}. International Journal of Applied Mathematics
  and Computer Science 27~(4), 697 -- 712.

\bibitem[{Bilalli et~al.(2018)Bilalli, Abell{\'{o}}, Aluja-Banet, and
  Wrembel}]{Bilalli2018}
Bilalli, B., Abell{\'{o}}, A., Aluja-Banet, T., Wrembel, R., 2018. Intelligent
  assistance for data pre-processing. Computer Standards and Interfaces 57, 101
  -- 109.

\bibitem[{Brazdil et~al.(1994)Brazdil, Gama, and Henery}]{Brazdil1994}
Brazdil, P., Gama, J., Henery, B., 1994. Characterizing the applicability of
  classification algorithms using meta-level learning. In: 7th European
  Conference on Machine Learning (ECML). pp. 83 -- 102.

\bibitem[{Brazdil et~al.(2009)Brazdil, Giraud-Carrier, Soares, and
  Vilalta}]{Brazdil2009}
Brazdil, P., Giraud-Carrier, C., Soares, C., Vilalta, R., 2009. Metalearning:
  Applications to Data Mining. Springer-Verlag Berlin Heidelberg.

\bibitem[{Brazdil et~al.(2003)Brazdil, Soares, and {da Coasta}}]{Brazdil2003}
Brazdil, P.~B., Soares, C., {da Coasta}, J.~P., 2003. Ranking learning
  algorithms: Using ibl and meta-learning on accuracy and time results. Machine
  Learning 50~(3), 251--277.

\bibitem[{Breiman et~al.(1984)Breiman, Friedman, Olshen, and
  Stone}]{Breiman1984}
Breiman, L., Friedman, J., Olshen, R.~A., Stone, C.~J., 1984. Classification
  and Regression Trees. Chapman and Hall.

\bibitem[{Burton et~al.(2014)Burton, Morris, Giraud-Carrier, West, and
  Thackeray}]{Burton2014}
Burton, S.~H., Morris, R.~G., Giraud-Carrier, C., West, J.~H., Thackeray, R.,
  2014. Mining useful association rules from questionnaire data. Intelligent
  Data Analysis 18~(3), 479--494.

\bibitem[{Castiello et~al.(2005)Castiello, Castellano, and
  Fanelli}]{Castiello2005}
Castiello, C., Castellano, G., Fanelli, A.~M., 2005. Meta-data:
  {C}haracterization of input features for meta-learning. In: 2nd International
  Conference on Modeling Decisions for Artificial Intelligence (MDAI). pp. 457
  -- 468.

\bibitem[{Engels and Theusinger(1998)}]{Engels1998}
Engels, R., Theusinger, C., 1998. Using a data metric for preprocessing advice
  for data mining applications. In: 13th European Conference on on Artificial
  Intelligence (ECAI). pp. 430 -- 434.

\bibitem[{Fayyad and Irani(1993)}]{Fayyad1993}
Fayyad, U.~M., Irani, K.~B., 1993. Multi-interval discretization of
  continuous-valued attributes for classification learning. In: 13th
  International Joint Conference on Artificial Intelligence (IJCAI). pp.
  1022--1029.

\bibitem[{Ferrari and de~Castro(2015)}]{Ferrari2015}
Ferrari, D.~G., de~Castro, L.~N., 2015. Clustering algorithm selection by
  meta-learning systems: {A} new distance-based problem characterization and
  ranking combination methods. Information Sciences 301, 181--194.

\bibitem[{Feurer et~al.(2014)Feurer, Springenberg, and Hutter}]{Feurer2014}
Feurer, M., Springenberg, J.~T., Hutter, F., 2014. Using meta-learning to
  initialize bayesian optimization of hyperparameters. In: International
  Conference on Meta-learning and Algorithm Selection (MLAS). pp. 3 -- 10.

\bibitem[{Filchenkov and Pendryak(2015)}]{Filchenkov2015}
Filchenkov, A., Pendryak, A., 2015. Datasets meta-feature description for
  recommending feature selection algorithm. In: Artificial Intelligence and
  Natural Language and Information Extraction, Social Media and Web Search
  FRUCT Conference (AINL-ISMW FRUCT). pp. 11 -- 18.

\bibitem[{F{\"{u}}rnkranz and Petrak(2001)}]{Furnkranz2001}
F{\"{u}}rnkranz, J., Petrak, J., 2001. An evaluation of landmarking variants.
  In: 1st ECML/PKDD International Workshop on Integration and Collaboration
  Aspects of Data Mining, Decision Support and Meta-Learning (IDDM). pp. 57 --
  68.

\bibitem[{Garcia et~al.(2015)Garcia, de~Carvalho, and Lorena}]{Garcia2015}
Garcia, L. P.~F., de~Carvalho, A. C. P. L.~F., Lorena, A.~C., 2015. Noise
  detection in the meta-learning level. Neurocomputing 176~(2), 1 -- 12.

\bibitem[{Garcia et~al.(2018)Garcia, Lorena, de~Souto, and Ho}]{Garcia2018}
Garcia, L. P.~F., Lorena, A.~C., de~Souto, M. C.~P., Ho, T.~K., 2018.
  Classifier recommendation using data complexity measures. In: 24th
  International Conference on Pattern Recognition, {ICPR}. pp. 874--879.

\bibitem[{Hall et~al.(2009)Hall, Frank, Holmes, Pfahringer, Reutemann, and
  Witten}]{Hall2009}
Hall, M., Frank, E., Holmes, G., Pfahringer, B., Reutemann, P., Witten, I.~H.,
  2009. The {WEKA} data mining software: An update. ACM SIGKDD Explorations
  Newsletter 11~(1), 10 -- 18.

\bibitem[{Han et~al.(2005)Han, Kamber, and Pei}]{Han2005}
Han, J., Kamber, M., Pei, J., 2005. Data Mining: Concepts and Techniques.
  Morgan Kaufmann.

\bibitem[{Handl et~al.(2005)Handl, Knowles, and Kell}]{Handl2005}
Handl, J., Knowles, J.~D., Kell, D.~B., 2005. Computational cluster validation
  in post-genomic data analysis. Bioinformatics 21~(15), 3201--3212.

\bibitem[{Ho and Basu(2002)}]{Ho2002}
Ho, T.~K., Basu, M., 2002. Complexity measures of supervised classification
  problems. IEEE Transactions on Pattern Analysis and Machine Intelligence
  24~(3), 289--300.

\bibitem[{Hotelling(1933)}]{Hotelling1933}
Hotelling, H., 1933. Analysis of a complex of statistical variables with
  principal components. Journal of Educational Psychology 24, 417--441.

\bibitem[{Hutson(2018)}]{Hutson2018}
Hutson, M., 2018. Artificial intelligence faces reproducibility crisis. Science
  359~(6377), 725 -- 726.

\bibitem[{Jin et~al.(2007)Jin, Breitbart, and Muoh}]{Jin2007}
Jin, R., Breitbart, Y., Muoh, C., 2007. Data discretization unification. In:
  7th International Conference on Data Mining (ICDM). pp. 183--192.

\bibitem[{Joanes and Gill(1998)}]{Joanes1998}
Joanes, D.~N., Gill, C.~A., 1998. Comparing measures of sample skewness and
  kurtosis. Journal of the Royal Statistical Society 47~(1), 183 -- 189.

\bibitem[{Kalousis(2002)}]{Kalousis2002}
Kalousis, A., 2002. Algorithm selection via meta-learning. Ph.D. thesis,
  Faculty of Science of the University of Geneva.

\bibitem[{Kalousis and Hilario(2001{\natexlab{a}})}]{Kalousis2001}
Kalousis, A., Hilario, M., 2001{\natexlab{a}}. Feature selection for
  meta-learning. In: 5th Pacific-Asia Conference on Knowledge Discovery and
  Data Mining (PAKDD). Vol. 2035. pp. 222--233.

\bibitem[{Kalousis and Hilario(2001{\natexlab{b}})}]{Kalousis2001a}
Kalousis, A., Hilario, M., 2001{\natexlab{b}}. Model selection via
  meta-learning: a comparative study. International Journal on Artificial
  Intelligence Tools 10~(4), 525 -- 554.

\bibitem[{Kalousis and Theoharis(1999)}]{Kalousis1999}
Kalousis, A., Theoharis, T., 1999. {NOEMON:} {D}esign, implementation and
  performance results of an intelligent assistant for classifier selection.
  Intelligent Data Analysis 3~(5), 319 -- 337.

\bibitem[{Kopf and Iglezakis(2002)}]{Kopf2002}
Kopf, C., Iglezakis, I., 2002. Combination of task description strategies and
  case base properties for meta-learning. In: 2nd ECML/PKDD International
  Workshop on Integration and Collaboration Aspects of Data Mining, Decision
  Support and Meta-Learning (IDDM). pp. 65 -- 76.

\bibitem[{Kopf et~al.(2000)Kopf, Taylor, and Keller}]{Kopf2000}
Kopf, C., Taylor, C., Keller, J., 2000. {Meta-Analysis}: {F}rom data
  characterisation for meta-learning to meta-regression. In: PKDD Workshop on
  Data Mining, Decision Support,Meta-Learning and Inductive Logic Programming.
  pp. 15 -- 26.

\bibitem[{Kuba et~al.(2002)Kuba, Brazdil, Soares, and Woznica}]{Kuba2002}
Kuba, P., Brazdil, P., Soares, C., Woznica, A., 2002. Exploiting sampling and
  meta-learning for parameter setting for support vector machines. In: 8th
  IBERAMIA Workshop on Learning and Data Mining. pp. 209 -- 216.

\bibitem[{Lee and Giraud-Carrier(2008)}]{Lee2008}
Lee, J.~W., Giraud-Carrier, C., 2008. Predicting algorithm accuracy with a
  small set of effective meta-features. In: 7th International Conference on
  Machine Learning and Applications (ICMLA). pp. 808--812.

\bibitem[{Leite and Brazdil(2005)}]{Leite2005}
Leite, R., Brazdil, P., 2005. Predicting relative performance of classifiers
  from samples. In: 22nd International Conference on Machine Learning (ICML).
  Vol. 119. pp. 497--503.

\bibitem[{Lemke et~al.(2015)Lemke, Budka, and Gabrys}]{Lemke2015}
Lemke, C., Budka, M., Gabrys, B., 2015. Metalearning: a survey of trends and
  technologies. Artificial Intelligence Review 44~(1), 117 -- 130.

\bibitem[{Ler et~al.(2018)Ler, Teng, He, and Gidijala}]{Ler2018}
Ler, D., Teng, H., He, Y., Gidijala, R., 2018. Algorithm selection for
  classification problems via cluster-based meta-features. In: {IEEE}
  International Conference on Big Data (Big Data). pp. 4952--4960.

\bibitem[{Lindner and Studer(1999)}]{Lindner1999}
Lindner, G., Studer, R., 1999. {AST}: {S}upport for algorithm selection with a
  {CBR} approach. In: European Conference on Principles of Data Mining and
  Knowledge Discovery (PKDD). pp. 418 -- 423.

\bibitem[{Loh(2014)}]{Loh2014}
Loh, W.-Y., 2014. Fifty years of classification and regression trees.
  International Statistical Review 82~(3), 329 -- 348.

\bibitem[{Lorena et~al.(2018)Lorena, Garcia, Lehmann, de~Souto, and
  Ho}]{Lorena2018}
Lorena, A.~C., Garcia, L. P.~F., Lehmann, J., de~Souto, M. C.~P., Ho, T.~K.,
  2018. How complex is your classification problem? {A} survey on measuring
  classification complexity. CoRR abs/1808.03591.

\bibitem[{Luengo and Herrera(2015)}]{Luengo2015}
Luengo, J., Herrera, F., 2015. An automatic extraction method of the domains of
  competence for learning classifiers using data complexity measures. Knowledge
  and Information Systems 42~(1), 147--180.

\bibitem[{Mathworks(2001)}]{mathworks2001}
Mathworks, 2001. Statistics toolbox: for use with {MATLAB}: user's guide.

\bibitem[{Michie et~al.(1994)Michie, Spiegelhalter, and Taylor}]{Michie1994}
Michie, D., Spiegelhalter, D.~J., Taylor, C.~C., 1994. Machine Learning, Neural
  and Statistical Classification. Ellis Horwood.

\bibitem[{Mierswa et~al.(2006)Mierswa, Wurst, Klinkenberg, Scholz, and
  Euler}]{Mierswa2006}
Mierswa, I., Wurst, M., Klinkenberg, R., Scholz, M., Euler, T., 2006. {YALE}:
  rapid prototyping for complex data mining tasks. In: 12th International
  Conference on Knowledge Discovery and Data Mining (KDD). pp. 935 -- 940.

\bibitem[{Mitchell(1997)}]{Mitchell1997}
Mitchell, T.~M., 1997. Machine Learning. McGraw Hill.

\bibitem[{Morais and Prati(2013)}]{Morais2013}
Morais, G., Prati, R.~C., 2013. Complex network measures for data set
  characterization. In: Brazilian Conference on Intelligent Systems (BRACIS).
  pp. 12--18.

\bibitem[{Mu{\~{n}}oz et~al.(2018)Mu{\~{n}}oz, Villanova, Baatar, and
  Smith{-}Miles}]{Munoz2018}
Mu{\~{n}}oz, M.~A., Villanova, L., Baatar, D., Smith{-}Miles, K., 2018.
  Instance spaces for machine learning classification. Machine Learning
  107~(1), 109--147.

\bibitem[{Nascimento et~al.(2009)Nascimento, Prud{\^{e}}ncio, de~Souto, and
  Costa}]{Nascimento2009}
Nascimento, A. C.~A., Prud{\^{e}}ncio, R. B.~C., de~Souto, M. C.~P., Costa,
  I.~G., 2009. Mining rules for the automatic selection process of clustering
  methods applied to cancer gene expression data. In: 19th International
  Conference on Artificial Neural Networks (ICANN). Vol. 5769. pp. 20--29.

\bibitem[{Nguyen et~al.(2012)Nguyen, Wang, Hilario, and Kalousis}]{Nguyen2012}
Nguyen, P., Wang, J., Hilario, M., Kalousis, A., 2012. Learning heterogeneous
  similarity measures for hybrid-recommendations in meta-mining. In: {IEEE}
  International Conference on Data Mining (ICDM). pp. 1026--1031.

\bibitem[{Peng et~al.(2002{\natexlab{a}})Peng, Flach, Brazdil, and
  Soares}]{Peng2002a}
Peng, Y., Flach, P.~A., Brazdil, P., Soares, C., 2002{\natexlab{a}}. Decision
  tree-based data characterization for meta-learning. In: 2nd ECML/PKDD
  International Workshop on Integration and Collaboration Aspects of Data
  Mining, Decision Support and Meta-Learning (IDDM). pp. 111 -- 122.

\bibitem[{Peng et~al.(2002{\natexlab{b}})Peng, Flach, Soares, and
  Brazdil}]{Peng2002}
Peng, Y., Flach, P.~A., Soares, C., Brazdil, P., 2002{\natexlab{b}}. Improved
  dataset characterisation for meta-learning. In: 5th International Conference
  on Discovery Science (DS). pp. 141 -- 152.

\bibitem[{Pfahringer et~al.(2000)Pfahringer, Bensusan, and
  Giraud-Carrier}]{Pfahringer2000}
Pfahringer, B., Bensusan, H., Giraud-Carrier, C., 2000. Meta-learning by
  landmarking various learning algorithms. In: 17th International Conference on
  Machine Learning (ICML). pp. 743 -- 750.

\bibitem[{Pimentel and de~Carvalho(2019)}]{Pimentel2019}
Pimentel, B.~A., de~Carvalho, A. C. P. L.~F., 2019. A new data characterization
  for selecting clustering algorithms using meta-learning. Information Sciences
  477, 203--219.

\bibitem[{Pinto et~al.(2016)Pinto, Soares, and Mendes-Moreira}]{Pinto2016}
Pinto, F., Soares, C., Mendes-Moreira, J., 2016. Towards automatic generation
  of metafeatures. In: Pacific-Asia Conference on Knowledge Discovery and Data
  Mining (PAKDD). pp. 215 -- 226.

\bibitem[{Reif(2012)}]{Reif2012}
Reif, M., 2012. A comprehensive dataset for evaluating approaches of various
  meta-learning tasks. In: 1st International Conference on Pattern Recognition
  Applications and Methods (ICPRAM). pp. 273 -- 276.

\bibitem[{Reif et~al.(2011)Reif, Shafait, and Dengel}]{Reif2011}
Reif, M., Shafait, F., Dengel, A., 2011. Prediction of classifier training time
  including parameter optimization. In: 34th German conference on Advances in
  artificial intelligence (KI). pp. 260 -- 271.

\bibitem[{Reif et~al.(2012)Reif, Shafait, and Dengel}]{Reif2012b}
Reif, M., Shafait, F., Dengel, A., 2012. {Meta$^2$-Features}: {P}roviding
  meta-learners more information. In: 35th German Conference on Artificial
  Intelligence (KI). pp. 74 -- 77.

\bibitem[{Reif et~al.(2014)Reif, Shafait, Goldstein, Breuel, and
  Dengel}]{Reif2014}
Reif, M., Shafait, F., Goldstein, M., Breuel, T., Dengel, A., 2014. Automatic
  classifier selection for non-experts. Pattern Analysis and Applications
  17~(1), 83 -- 96.

\bibitem[{Rodgers and Nicewander(1988)}]{Rodgers1988}
Rodgers, J.~L., Nicewander, W.~A., 1988. Thirteen ways to look at the
  correlation coefficient. The American Statistician 42~(1), 59 -- 66.

\bibitem[{Rousseeuw and Hubert(2011)}]{Rousseeuw2011}
Rousseeuw, P.~J., Hubert, M., 2011. Robust statistics for outlier detection.
  Wiley Interdisciplinary Reviews: Data Mining and Knowledge Discovery 1~(1),
  73 -- 79.

\bibitem[{Royston(1995)}]{Royston1995}
Royston, P., 1995. Remark {AS R94}: {A} remark on algorithm {AS 181}: {T}he
  {W}-test for normality. Journal of the Royal Statistical Society. Series C
  (Applied Statistics) 44~(4), 547 -- 551.

\bibitem[{Salama et~al.(2013)Salama, Hassanien, and Revett}]{Salama2013}
Salama, M.~A., Hassanien, A.~E., Revett, K., 2013. Employment of neural network
  and rough set in meta-learning. Memetic Computing 5~(3), 165 -- 177.

\bibitem[{Segrera et~al.(2008)Segrera, Pinho, and Moreno}]{Segrera2008}
Segrera, S., Pinho, J., Moreno, M.~N., 2008. Information-theoretic measures for
  meta-learning. In: Hybrid Artificial Intelligence Systems (HAIS). pp.
  458--465.

\bibitem[{Smith et~al.(2001)Smith, Woo, Ciesielski, and Ibrahim}]{Smith2001}
Smith, K., Woo, F., Ciesielski, V., Ibrahim, R., 2001. Modelling the
  relationship between problem characteristics and data mining algorithm
  performance using neural networks. In: Smart Engineering System Design:
  Neural Networks, Fuzzy Logic, Evolutionary Programming, Data Mining, and
  Complex Systems. pp. 357 -- 362.

\bibitem[{Smith et~al.(2014)Smith, Martinez, and Giraud-Carrier}]{Smith2014}
Smith, M.~R., Martinez, T., Giraud-Carrier, C., 2014. An instance level
  analysis of data complexity. Machine Learning 95~(2), 225--256.

\bibitem[{Smith{-}Miles(2008)}]{Smith-Miles2008}
Smith{-}Miles, K., 2008. Cross-disciplinary perspectives on meta-learning for
  algorithm selection. {ACM} Computing Surveys 41~(1), 6:1--6:25.

\bibitem[{Soares et~al.(2001)Soares, Petrak, and Brazdil}]{Soares2001a}
Soares, C., Petrak, J., Brazdil, P., 2001. Sampling-based relative landmarks:
  Systematically test-driving algorithms before choosing. In: 10th Portuguese
  Conference on Artificial Intelligence (EPIA). pp. 88 -- 95.

\bibitem[{Sohn(1999)}]{Sohn1999}
Sohn, S.~Y., 1999. Meta analysis of classification algorithms for pattern
  recognition. IEEE Transactions on Pattern Analysis and Machine Intelligence
  21~(11), 1137 -- 1144.

\bibitem[{Song et~al.(2012)Song, Wang, and Wang}]{Song2012}
Song, Q., Wang, G., Wang, C., 2012. Automatic recommendation of classification
  algorithms based on data set characteristics. Pattern Recognition 45~(7),
  2672--2689.

\bibitem[{Sun and Pfahringer(2013)}]{Sun2013}
Sun, Q., Pfahringer, B., 2013. Pairwise meta-rules for better
  meta-learning-based algorithm ranking. Machine Learning 93~(1), 141--161.

\bibitem[{Tan et~al.(2005)Tan, Steinbach, and Kumar}]{Tan2005}
Tan, P.-N., Steinbach, M., Kumar, V., 2005. Introduction to Data Mining.
  Addison-Wesley Longman Publishing.

\bibitem[{Todorovski et~al.(2000)Todorovski, Brazdil, and
  Soares}]{Todorovski2000}
Todorovski, L., Brazdil, P., Soares, C., 2000. Report on the experiments with
  feature selection in meta-level learning. In: PKDD Workshop on Data Mining,
  Decision Support, Meta-Learning and Inductive Logic Programming. pp. 27 --
  39.

\bibitem[{Vanschoren(2010)}]{Vanschoren2010}
Vanschoren, J., 2010. Understanding machine learning performance with
  experiment databases. Ph.D. thesis, Leuven Univeristy.

\bibitem[{Vanschoren et~al.(2012)Vanschoren, Blockeel, Pfahringer, and
  Holmes}]{Vanschoren2012}
Vanschoren, J., Blockeel, H., Pfahringer, B., Holmes, G., 2012. Experiment
  databases. Machine Learning 87~(2), 127--158.

\bibitem[{Vanschoren et~al.(2013)Vanschoren, {van Rijn}, Bischl, and
  Torgo}]{OpenML2013}
Vanschoren, J., {van Rijn}, J.~N., Bischl, B., Torgo, L., 2013. {OpenML}:
  {N}etworked science in machine learning. ACM SIGKDD Explorations Newsletter
  15~(2), 49 -- 60.

\bibitem[{Vilalta(1999)}]{Vilalta1999}
Vilalta, R., 1999. Understanding accuracy performance through concept
  characterization and algorithm analysis. In: ECML Workshop on Recent Advances
  in Meta-Learning and Future Work. pp. 3--9.

\bibitem[{Vilalta and Drissi(2002)}]{Vilalta2002a}
Vilalta, R., Drissi, Y., 2002. A characterization of difficult problems in
  classification. In: International Conference on Machine Learning and
  Applications (ICMLA). pp. 133--138.

\bibitem[{Vukicevic et~al.(2016)Vukicevic, Radovanovic, Delibasic, and
  Suknovic}]{Vukicevic2016}
Vukicevic, M., Radovanovic, S., Delibasic, B., Suknovic, M., 2016. Extending
  meta-learning framework for clustering gene expression data with
  component-based algorithm design and internal evaluation measures.
  International Journal of Data Mining and Bioinformatics (IJDMB) 14~(2),
  101--119.

\bibitem[{Wang et~al.(2013)Wang, Song, Sun, Zhang, Xu, and Zhou}]{Wang2013}
Wang, G., Song, Q., Sun, H., Zhang, X., Xu, B., Zhou, Y., 2013. A feature
  subset selection algorithm automatic recommendation method. Journal of
  Artificial Intelligence Research 47, 1--34.

\bibitem[{Wang et~al.(2015)Wang, Song, and Zhu}]{Wang2015}
Wang, G., Song, Q., Zhu, X., 2015. An improved data characterization method and
  its application in classification algorithm recommendation. Applied
  Intelligence 43~(4), 892--912.

\bibitem[{Wolpert(1992)}]{Wolpert1992}
Wolpert, D.~H., 1992. Stacked generalization. Neural Networks 5~(2), 241 --
  259.

\end{thebibliography}





\end{document}